\documentclass{article} 
\usepackage[final]{colm2026_conference}

\usepackage{iftex}
\RequireXeTeX
\usepackage{fontspec}
\newfontfamily\thaifont{Norasi}[
  Path           = fonts/,
  Extension      = .otf,
  UprightFont    = *,
  BoldFont       = *-Bold,
  ItalicFont     = *-Italic,
  BoldItalicFont = *-BoldItalic,
  Script         = Thai,
  Scale          = MatchLowercase]

\usepackage{ucharclasses}
\makeatletter
\newcommand{\thaiSavedFamily}{\rmdefault}
\newcommand{\thaiEnter}{\xdef\thaiSavedFamily{\f@family}\thaifont
  \XeTeXlinebreaklocale "th"\XeTeXlinebreakskip=0pt plus 0.1pt\relax}
\newcommand{\thaiLeave}{\fontfamily{\thaiSavedFamily}\selectfont
  \XeTeXlinebreaklocale ""\XeTeXlinebreakskip=0pt\relax}
\makeatother
\setTransitionsFor{Thai}{\thaiEnter}{\thaiLeave}

\usepackage[table]{xcolor}
\usepackage{graphicx}
\usepackage{microtype}
\usepackage{hyperref}
\usepackage{url}
\usepackage{booktabs}
\usepackage{caption}
\usepackage{algorithm}
\usepackage{algpseudocode}
\usepackage[nobar]{wayupaxa}
\reportstamp{Technical Report}
\resources{%
  \resourcerow{Model}{\href{https://huggingface.co/wayu-ai/wayu-paxa-ocr-zero}{\texttt{wayu-ai/wayu-paxa-ocr-zero}}}
  \resourcerow{Reconstruction}{\href{https://github.com/wayu-research/docaug}{\texttt{wayu-research/docaug}}}}

\usepackage{lineno}

\definecolor{darkblue}{rgb}{0, 0, 0.5}
\hypersetup{colorlinks=true, citecolor=darkblue, linkcolor=darkblue, urlcolor=darkblue}

\title{How Far Can Synthetic Data Take Thai OCR?}

\author{Kunat Pipatanakul$^{1,2}$\\ \vspace{0.4em} {\normalfont\small $^{1}$Wayu Research \quad $^{2}$Paxa Labs} }

\begin{document}

\ifcolmsubmission
\linenumbers
\fi

\maketitle

\begin{abstract}
We investigate what makes synthetic OCR supervision transfer to real Thai documents and
use the resulting insights to build Wayu-Paxa-OCR-Zero, a Thai OCR model adapted without OCR
labels from real Thai document pages. Synthetic data provide exact labels at scale, but
``realism'' conflates source domain, page context, typography, spatial structure, and glyph
variation. We disentangle these factors with a controlled document-reconstruction pipeline
and evaluate each variant under page- and crop-level training on printed and handwritten
Thai documents. Non-text context has little consistent effect, whereas typeface diversity,
two-dimensional structure, and real handwriting glyphs improve transfer; moreover, source-
domain matching depends on training granularity, with in-domain reconstruction approaching
real printed supervision under page-level training (1.82\% versus 1.31\% median character
error rate) but underperforming out-of-domain reconstruction under crop-level training
(15.59\% versus 5.52\%). Guided by these findings, we adapt the 0.9B-parameter
PaddleOCR-VL-1.6 into Wayu-Paxa-OCR-Zero using 45,723 synthetic pages: relative to its base
checkpoint, it reduces
median character error rate from 6.64\% to 1.24\% on printed pages and from 74.87\% to
20.55\% on handwriting and outperforms Typhoon OCR v1 7B on all five evaluation sets,
showing that synthetic-only training can be competitive.
\end{abstract}

\section{Introduction}
\label{sec:introduction}

Optical character recognition (OCR) converts document images into machine-readable text for
digitization, search, and retrieval-augmented generation. Classical systems such as
Tesseract rely on specialized recognition pipelines~\citep{smith2007tesseract}, whereas
modern vision--language models (VLMs) perform end-to-end recognition while preserving
reading order and document context. Proprietary systems such as Gemini and GPT provide
strong multilingual OCR capabilities~\citep{geminiteam2023gemini,openai2024gpt4o}; open
models such as Unlimited OCR and PaddleOCR-VL enable lower-cost local processing of
sensitive documents~\citep{yin2026unlimitedocrworks,zhang2026paddleocrvl16}.

Coverage, however, remains uneven. Open models, datasets, and benchmarks center on English
and Chinese, while less-resourced languages often have abundant documents but few reliable
labels. Thai is our case study: its unique glyph system doesn't allow transfer from English or Chinese. While PDF text extraction and OCR pseudo-labels can omit characters, reorder
combining marks, and corrupt reading order. Manual correction is costly, and open Thai
datasets remain limited.

Typhoon OCR defines the open Thai frontier. Its 2B V1.5 model achieves state-of-the-art Thai
performance and competes with larger proprietary systems~\citep{nonesung2026typhoonocr}.
Its training pipeline combines traditional OCR, VLM restructuring, and curated synthetic
data. Typhoon OCR therefore establishes the value of Thai-specific adaptation, but does not
isolate the contribution of synthetic supervision or the document properties that support transfer.

Recent work demonstrates synthetic OCR transfer with layout-aware Indic
pages~\citep{kolavi2025nayana}, Manchu word images~\citep{chung2025manchu},
cross-lingual Arabic document reconstruction~\citep{alhomoud2025synthdocs}, and
degradation-aware historical pages~\citep{guan2025prepocr}. These approaches vary
several generation factors together, leaving it unclear whether transfer comes from
layout, non-text context, fonts, or training granularity. The last distinction
matters because whole-page models and modern detector--recognizer systems such as GLM-OCR
and PaddleOCR-VL expose different amounts of document
context~\citep{duan2026glmocr,zhang2026paddleocrvl16}.

To this end, we ask a central question: \emph{How far can synthetic data take Thai OCR?}
We answer it through controlled reconstruction of naturally occurring documents. Our
pipeline renders OCR labels into their source regions while varying the source domain,
non-text context, typeface distribution, two-dimensional layout, and handwriting glyph
source. Using Qwen3-VL-2B-Instruct~\citep{bai2025qwen3vl}, we compare page-level and crop-level
training on out-of-domain reconstructions of public English documents and in-domain
reconstructions of real Thai documents. We then compare reconstruction with real Thai
supervision. Based on these findings, we derive a synthetic training recipe and use it to
train \textbf{Wayu-Paxa-OCR-Zero}, a Thai adaptation of PaddleOCR-VL-1.6 trained only on
synthetic supervision.

We summarize our contributions as follows:
\begin{itemize}
  \item \textbf{Controllable document reconstruction.} We introduce a pipeline that
    replaces source text in place while independently controlling source domain, non-text
    context, typeface diversity, two-dimensional layout, and handwriting glyph source.
  \item \textbf{Evidence about synthetic-to-real transfer.} Controlled page- and crop-level
    experiments identify typography, spatial structure, glyph variation, and the interaction
    between source domain and training granularity as key determinants of transfer.
  \item \textbf{A synthetic-supervision Thai OCR model.} We introduce Wayu-Paxa-OCR-Zero,
    trained using synthetic data generated from 45,723 pages. The model substantially
    improves its base checkpoint and outperforms the Typhoon OCR 7B model on all five
    evaluation sets.
\end{itemize}

\section{Synthetic OCR from Reconstructed Documents}
\label{sec:synthetic-pipeline}

\begin{figure}[h!]
  \centering
  \includegraphics[width=\textwidth]{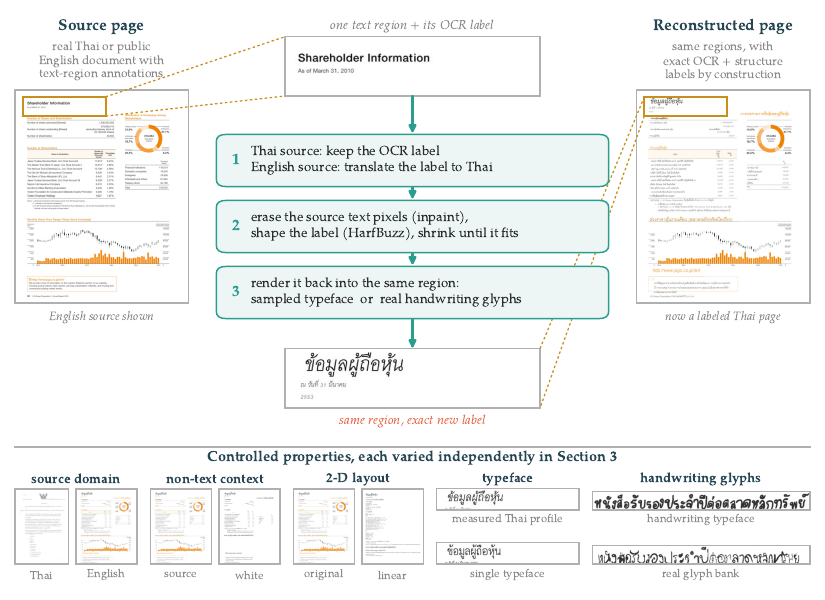}
  \caption{Overview of the document reconstruction pipeline. Thai sources provide OCR
  labels directly for In-Domain Reconstruction, while non-Thai sources are translated into
  Thai or retained in English for Out-of-Domain Reconstruction. The reconstruction settings
  control the retained page context and render each OCR label using either typefaces sampled
  from a specified distribution or real handwriting glyphs.}
  \label{fig:reconstruction-pipeline}
\end{figure}

We generate synthetic Thai OCR pages by reconstructing existing documents in place.
Figure~\ref{fig:reconstruction-pipeline} summarizes the pipeline. For Thai sources, we use
the OCR label of each text region. For non-Thai sources, we either translate the source text
into Thai or retain its English OCR label. We then erase / inpaint the source text pixels, fit the OCR
label to the original region, and render it with either a sampled typeface or real
handwriting glyphs. The reconstruction settings control the source domain, layout,
background, non-text page context, typeface distribution, and glyph source.

\subsection{In-Domain and Out-of-Domain Reconstruction}

Each source example contains a page image and annotated text regions. We retain the region
geometry, document-element categories, and reading order when available. For In-Domain
Synthetic, we use the OCR label of each Thai source region. For Out-of-Domain Synthetic, we
translate most non-Thai OCR labels into Thai, following the translation-based data
construction used by Typhoon and Typhoon 2~\citep{pipatanakul2023typhoon,pipatanakul2024typhoon2}.
We then inpaint the source text pixels and render the OCR label in the corresponding region.

The standard reconstruction retains the background and non-text pixels from the source
page. To control page context, we replace these pixels with a white background while keeping
the text regions fixed. To control layout, we retain the original two-dimensional
arrangement or stack the regions vertically.

\subsection{Fit-Constrained Rendering}

Thai translations need not match the length of their English sources, and Thai vowels and
tone marks can occupy multiple vertical levels. We therefore shape the complete
region-level OCR label with HarfBuzz before placement. We sample the typeface and type size
independently of the source text, then reduce the type size until the complete OCR label fits
the original region. If the OCR label still overflows at the minimum acceptable size, we
reject the complete page.

\subsection{Typeface Rendering}

For typeface rendering, we sample a Thai typeface for each shaped OCR label from a specified
distribution. We use the measured profile in Section~\ref{sec:font-glyph} by default and
replace it with a single typeface in the controlled experiment. The distribution includes
both printed and handwriting typefaces. Repeated occurrences of a character rendered with
the same typeface share the same outline.

\subsection{Handwriting Real-Glyph Rendering}

For handwriting real-glyph rendering, we replace supported Thai characters with instances sampled from the handwriting banks in Section~\ref{sec:font-glyph}. We sample each character
independently, so repeated characters can use different strokes. Unsupported characters
fall back to typeface rendering. We keep the OCR label, region annotations, ink height,
and ink color fixed between the typeface and real-glyph renderings.

Section~\ref{sec:transfer} uses these reconstruction settings to study how each controlled
property affects transfer to real Thai documents.

\section{What Makes Synthetic Data Transfer to Real Thai Documents?}
\label{sec:transfer}

We use the reconstruction controls from Section~\ref{sec:synthetic-pipeline} to study which
properties transfer to real Thai documents. We first vary non-text page context, typeface
diversity, and two-dimensional layout within Out-of-Domain Synthetic. We then compare source
domains, page-level and crop-level training, synthetic and real supervision, and typeface
and real-glyph handwriting.

\subsection{Experimental Setup}

\paragraph{Models and training settings.}

We use Qwen3-VL-2B-Instruct as the main experimental model and compare two training
settings~\citep{bai2025qwen3vl}:
\begin{itemize}
  \item \textbf{Page-level training:} The model receives a complete document image and
    predicts the full page and its regions in a single inference pass.
    Appendix~\ref{app:formats} gives the instruction and the target schema.
  \item \textbf{Crop-level training:} The model recognizes individual regions produced by
    a layout detector~\citep{sun2025ppdoclayout}. This setting follows two-stage systems
    such as GLM-OCR and PaddleOCR-VL-1.6, which use PP-DocLayoutV3 before VLM
    recognition~\citep{duan2026glmocr,zhang2026paddleocrvl16}.
    Appendix~\ref{app:formats} gives the prediction formats.
\end{itemize}

\paragraph{Data sources.}

Training data in this study primarily target Thai OCR. We group the data by how their OCR
labels are obtained.
\begin{itemize}
  \item \textbf{Real Thai (Print):} We gather real Thai documents from public Thai PDFs and
    document images from Common Crawl~\citep{commoncrawl2026} and other websites, including
    government documents, forms, scans, reports, and online publications. The collection
    contains approximately 34,000 pages and is split into training pages and a test set. We use the
    training split in two ways: 1) as source documents for In-Domain Synthetic, where the
    original text pixels are inpainted and replaced by rendered OCR labels in
    Section~\ref{sec:in-domain}, and 2) with the original OCR labels as real printed
    supervision in Section~\ref{sec:real-supervision}. The test split forms the Heldout
    evaluation set described below.
  \item \textbf{Real Thai (Handwriting):} We gather photographed Thai study-notebook
    pages from public websites. The collection contains approximately 4,000 pages and is split into training pages and 400 test pages. In Section~\ref{sec:real-supervision},
    we add the training split to Real Thai (Print) to form the Real Thai (Print +
    Handwriting) condition; Section~\ref{sec:handwriting} includes this condition as a
    reference. The held-out pages form the Handwriting and Easy Handwriting evaluation sets
    described below.
  \item \textbf{Out-of-Domain Synthetic:} This source, used in our main training experiment,
    simulates a setting in which Thai documents are unavailable and only public English
    datasets are accessible. Specifically, we apply the pipeline in
    Section~\ref{sec:synthetic-pipeline} to English pages from
    DocLayNet~\citep{Pfitzmann_2022}, designed pages from
    Crello~\citep{yamaguchi2021canvasvae}, and wide tables from
    PubTabNet~\citep{zhong2020pubtabnet}. We translate most source text into Thai and render
    the resulting OCR labels in the original regions while retaining the source layout and
    non-text pixels. The remaining 7.61\% of pages retain their English OCR labels,
    approximating the English-language proportion in Real Thai (Print). This
    dataset and its variants are used in
    Sections~\ref{sec:source-factors}--\ref{sec:handwriting}.
  \item \textbf{In-Domain Synthetic:} For this source, we apply the pipeline in
    Section~\ref{sec:synthetic-pipeline} to reconstruct the Real Thai (Print) pages with
    their OCR labels. This source tests whether transfer benefits from real Thai non-text
    context, document layout, and typographic style. We use this dataset in
    Sections~\ref{sec:in-domain}--\ref{sec:real-supervision}.
\end{itemize}
Appendix~\ref{app:real-labels} describes how we construct the OCR labels for Real Thai
(Print) and Real Thai (Handwriting).

\paragraph{Font \& Glyph.}
\label{sec:font-glyph}

We instantiate text appearance using two complementary sources:
\begin{itemize}
  \item \textbf{Fonts:} We shape Thai text with HarfBuzz and fit the type size to each region.
    Typeface sampling follows a character-weighted profile measured from 8,000 public Thai
    PDF pages. Table~\ref{tab:thai-font-distribution} summarizes
    the measured distribution.
  \item \textbf{Real glyphs:} We construct a handwriting glyph bank from the
    iApp Handwriting Dataset~\citep{iapp2024thaihandwriting} and the Real Thai (Handwriting)
    training split using the pipeline in Appendix~\ref{app:glyph-bank}. The bank contains approximately 6,000 instances across
    76 character classes, covering Thai consonants, vowels, tone marks, and digits.
    Unsupported characters fall back to typeface rendering.
\end{itemize}

\begin{table}[h!]
  \centering
  \caption{Character-weighted font-family distribution measured from 8,000 randomly
  sampled pages. The profile contains 693 observed family names; the ten most frequent
  account for 80.9\% of Thai characters.}
  \label{tab:thai-font-distribution}
  \small
  \begin{tabular}{lr}
    \toprule
    Font family & Thai characters (\%) \\
    \midrule
    Cordia New       & 18.6 \\
    Angsana New      & 15.3 \\
    TH Sarabun PSK   & 14.7 \\
    TH Sarabun New   & 14.2 \\
    Browallia New    & 6.9 \\
    TH Sarabun IT๙   & 4.7 \\
    DB ThaiText X    & 2.4 \\
    Cordia UPC       & 1.9 \\
    Tahoma           & 1.1 \\
    DB FongNam X     & 1.1 \\
    Other (683)      & 19.1 \\
    \bottomrule
  \end{tabular}
\end{table}

\paragraph{Evaluation data and metrics.}

We evaluate on three datasets:
\begin{itemize}
  \item \textbf{Heldout:} 301 real printed pages from the test split of Real Thai (Print),
    disjoint from its training split.
  \item \textbf{Handwriting:} 200 photographed Thai study-notebook pages from the
    held-out portion of Real Thai (Handwriting), one per writer and disjoint from training
    by page and writer identity. The set spans a broad range of legibility.
  \item \textbf{Easy Handwriting:} 200 pages from the same held-out handwriting population,
    restricted to the top of the legibility band based on low disagreement between two
    proprietary handwriting recognition systems.
\end{itemize}
We construct the reference OCR labels for all three evaluation sets using the evaluation
pipeline in Appendix~\ref{app:real-labels}.
We report character error rate (CER; lower is better) over text-only regions using two aggregates:
1) \textbf{Median} is the median page CER, and 2) \textbf{Mean} is total edit distance
divided by total reference characters.
These aggregates characterize complementary behavior: the median reflects performance on
a typical page and is less sensitive to severe failures, whereas the mean measures
aggregate error across all reference characters and weights pages by length. All CER
values are given in percent. To compare page-level and crop-level systems, we use fuzzy
alignment to project each prediction onto the evaluation regions before scoring
(Appendix~\ref{app:projected-cer}), as document-parsing benchmarks match predicted blocks to
reference blocks before scoring~\citep{ouyang2025omnidocbench,li2025readoc}. Following
the contract-dependent ignore handling of OmniDocBench, this
projection removes non-target page elements while retaining errors and missing regions
within the evaluated regions. We evaluate the element types specified by OmniDocBench.

\paragraph{Training parameters.}

Unless otherwise stated, we train all models for one epoch using
AdamW~\citep{loshchilov2019decoupled}. We use a learning rate of $3\times10^{-5}$ with a
cosine schedule and update all model parameters.

\subsection{Which Source-Document Properties Matter?}
\label{sec:source-factors}

This experiment studies which synthetic components affect recognition. Specifically,
\textbf{White-layout} renders the original text layout on a white background, retaining the
text regions while removing backgrounds, figures, rules, and scan artifacts.
\textbf{White-single-font} additionally replaces the font distribution with one typeface,
and \textbf{Linear-white-single-font} removes the layout component by stacking the regions
vertically instead of preserving their two-dimensional arrangement. We train
each variant using both page-level and crop-level training.
Figure~\ref{fig:ablation-examples} in Appendix~\ref{app:ablation-examples} shows two pages
rendered under all four variants.

\begin{table}[h!]
  \centering
  \caption{Source-property ablation. Values are CER in percent (lower is better); Med.\ is
  the page median and Mean is the character-weighted CER.}
  \label{tab:transfer-ablation}
  \small
  \setlength{\tabcolsep}{5pt}
  \begin{tabular}{llcccccc}
    \toprule
    & & \multicolumn{2}{c}{Heldout} & \multicolumn{2}{c}{Handwriting}
      & \multicolumn{2}{c}{Easy Handwriting} \\
    \cmidrule(lr){3-4} \cmidrule(lr){5-6} \cmidrule(lr){7-8}
    Training data & Regime & Med. & Mean & Med. & Mean & Med. & Mean \\
    \midrule
    Out-of-Domain Synthetic & page & 5.07 & 16.90 & 43.99 & 46.71 & 38.90 & 44.21 \\
    $-$ non-text context        & page & \textbf{4.78} & 21.05 & \textbf{43.16}
      & \textbf{44.01} & \textbf{38.07} & \textbf{41.38} \\
    $-$ font diversity          & page & 5.34 & 20.37 & 49.86 & 50.88 & 47.55 & 50.80 \\
    $-$ two-dimensional layout  & page & 5.07 & \textbf{13.58} & 58.40 & 59.97 & 49.60 & 53.92 \\
    \midrule
    Out-of-Domain Synthetic & crop & 5.52 & 19.09 & \textbf{49.15} & 49.37 & 48.26 & 49.50 \\
    $-$ non-text context        & crop & \textbf{4.86} & 16.69 & 50.77 & \textbf{49.03}
      & \textbf{47.81} & \textbf{49.08} \\
    $-$ font diversity          & crop & 7.01 & \textbf{16.51} & 62.13 & 59.92 & 61.18 & 61.28 \\
    $-$ two-dimensional layout  & crop & 9.60 & 20.69 & 70.17 & 67.02 & 67.19 & 65.64 \\
    \midrule
    Qwen3-VL-2B-Instruct    & page & 14.47 & 28.62 & 61.59 & 61.25 & 59.29 & 61.32 \\
    \bottomrule
  \end{tabular}
\end{table}

Removing non-text context has no consistent effect on recognition: median CER changes by at
most 1.62 points across the three datasets and two training settings. Removing font diversity
produces the first consistent loss on handwriting. Median CER increases by 6.70 and 9.48
points under page-level training and by 11.36 and 13.37 points under crop-level training,
while the increase on printed Heldout is 0.56 and 2.15 points. Typography diversity therefore
matters most when the target appearance extends beyond printed text.

Flattening the remaining layout further increases handwriting CER in both training settings.
For crop-level training, it also raises the Heldout median from 7.01 to 9.60; the page-level
median remains nearly unchanged at 5.07. Across the variants, printed page-level recognition
is stable, whereas handwriting degrades monotonically once font diversity and
two-dimensional structure are removed. Non-text context provides little CER benefit, while
typography and spatial structure improve recognition on the out-of-distribution handwriting
sets.

\subsection{Does In-Domain Reconstruction Help?}
\label{sec:in-domain}

The source-property ablation uses the same English source documents. We next
compare Out-of-Domain Synthetic with In-Domain Synthetic, which applies the same
reconstruction pipeline to Real Thai (Print). The original text is erased
and its OCR label is rendered into the same regions. The evaluation pages are disjoint from
these source pages. This comparison changes the domain of the source documents. The
resulting pages retain the layout, writing style, font distribution, and noise patterns of
Thai source documents.
Figure~\ref{fig:domain-examples} in Appendix~\ref{app:domain-examples} shows the two
synthetic sources beside the real Thai pages the in-domain one is built from.

\begin{table}[h!]
  \centering
  \caption{Comparison of Out-of-Domain Synthetic and In-Domain Synthetic. Values are CER
  in percent (lower is better); Med.\ is the page median and
  Mean is the character-weighted CER. The better result within each training block is
  bold.}
  \label{tab:indomain}
  \small
  \setlength{\tabcolsep}{5pt}
  \begin{tabular}{llcccccc}
    \toprule
    & & \multicolumn{2}{c}{Heldout} & \multicolumn{2}{c}{Handwriting}
      & \multicolumn{2}{c}{Easy Handwriting} \\
    \cmidrule(lr){3-4} \cmidrule(lr){5-6} \cmidrule(lr){7-8}
    Training data & Regime & Med. & Mean & Med. & Mean & Med. & Mean \\
    \midrule
    Out-of-Domain Synthetic & page & 5.07 & 16.90 & 43.99 & 46.71 & 38.90 & 44.21 \\
    In-Domain Synthetic     & page & \textbf{1.82} & \textbf{16.20} & \textbf{36.27}
      & \textbf{38.20} & \textbf{38.77} & \textbf{41.12} \\
    \midrule
    Out-of-Domain Synthetic & crop & \textbf{5.52} & \textbf{19.09} & \textbf{49.15}
      & \textbf{49.37} & \textbf{48.26} & \textbf{49.50} \\
    In-Domain Synthetic     & crop & 15.59 & 27.70 & 52.77 & 51.35 & 51.71 & 52.40 \\
    \midrule
    Qwen3-VL-2B-Instruct     & page & 14.47 & 28.62 & 61.59 & 61.25 & 59.29 & 61.32 \\
    \bottomrule
  \end{tabular}
\end{table}

Under page-level training, replacing Out-of-Domain Synthetic with In-Domain Synthetic
reduces median CER from 5.07 to 1.82 on Heldout and from 43.99 to 36.27 on Handwriting.

The result reverses under crop-level training. Replacing Out-of-Domain Synthetic with
In-Domain Synthetic increases median CER from 5.52 to 15.59 on Heldout, from 49.15 to 52.77
on Handwriting, and from 48.26 to 51.71 on Easy Handwriting. In-domain reconstruction
therefore helps the page-level model but not the crop-level model in this comparison.
Section~\ref{sec:granularity} summarizes the difference between page-level and crop-level
behavior.

\subsection{Does the Training Granularity Determine What Transfers?}
\label{sec:granularity}

Tables~\ref{tab:transfer-ablation} and~\ref{tab:indomain} evaluate each synthetic dataset
with both page-level and crop-level training. These settings produce different systems:
the page model observes the complete document, whereas the crop pipeline observes only an
individual element. We therefore ask separately whether the training unit changes the
conclusions and whether it changes the preferred dataset.

In summary, under both settings, removing non-text context has a
small and inconsistent effect, while removing font diversity and two-dimensional structure
progressively degrades both handwriting sets. The preferred source domain, however, changes
with the training unit. In-Domain Synthetic gives the lowest Heldout CER among the synthetic
datasets under page-level training (1.82), but performs substantially worse than
Out-of-Domain Synthetic under crop-level training (15.59 versus 5.52). The cause of this
reversal remains unclear and warrants further study.

\subsection{How Close Can Reconstruction Get to Real Supervision?}
\label{sec:real-supervision}

Table~\ref{tab:real-bound} compares the two synthetic training sets with training on Real
Thai (Print). All results in this comparison use page-level training.

\begin{table}[h!]
  \centering
  \caption{Reconstruction against real supervision under page-level training. Values are
  CER in percent (lower is better); Med.\ is the page median and
  Mean is the character-weighted CER. The last row combines Real Thai (Print) and Real Thai
  (Handwriting); the best result in each column is bold.}
  \label{tab:real-bound}
  \small
  \setlength{\tabcolsep}{5pt}
  \begin{tabular}{lcccccc}
    \toprule
    & \multicolumn{2}{c}{Heldout} & \multicolumn{2}{c}{Handwriting}
      & \multicolumn{2}{c}{Easy Handwriting} \\
    \cmidrule(lr){2-3} \cmidrule(lr){4-5} \cmidrule(lr){6-7}
    Training data & Med. & Mean & Med. & Mean & Med. & Mean \\
    \midrule
    Qwen3-VL-2B-Instruct    & 14.47 & 28.62 & 61.59 & 61.25 & 59.29 & 61.32 \\
    \midrule
    Out-of-Domain Synthetic & 5.07 & 16.90 & 43.99 & 46.71 & 38.90 & 44.21 \\
    In-Domain Synthetic     & 1.82 & 16.20 & 36.27 & 38.20 & 38.77 & 41.12 \\
    \midrule
    Real Thai (Print)           & \textbf{1.31} & 9.79 & 36.14 & 39.05 & 32.98 & 36.95 \\
    Real Thai (Print + Handwriting) & 1.40 & \textbf{8.49} & \textbf{26.05} & \textbf{28.26} & \textbf{20.79}
      & \textbf{25.55} \\
    \bottomrule
  \end{tabular}
\end{table}

On printed Heldout pages, In-Domain Synthetic approaches real supervision on the typical
page: its median CER of 1.82\% is only 0.51 points above Real Thai (Print) at 1.31\%.
The gap is substantially larger under mean CER, however, with 16.20\% for In-Domain
Synthetic versus 9.79\% for Real Thai (Print). Reconstruction therefore captures much of
what is needed for typical printed-page recognition, but real supervision still reduces a
tail of severe errors that disproportionately affects the character-weighted aggregate.
Out-of-Domain Synthetic remains further behind at 5.07\% median CER, showing that matching
the source-document domain further narrows the synthetic-to-real gap.

The handwriting results reveal a different limitation. In-Domain Synthetic and Real Thai
(Print) perform nearly identically on the broader Handwriting set (36.27\% versus 36.14\%
median CER), indicating that reconstructing real Thai printed pages recovers most of the
handwriting transfer obtained from real printed supervision. Neither condition, however,
approaches training with real handwriting: adding Real Thai (Handwriting) reduces median
CER to 26.05\% on Handwriting and 20.79\% on Easy Handwriting. The remaining gap is
therefore not explained by document domain alone; it points to appearance variation in
real handwriting that printed reconstruction does not capture.

Taken together, reconstruction comes close to real supervision for printed Thai,
particularly on typical pages, but does not fully reproduce the robustness or handwriting
variation provided by real data. Section~\ref{sec:handwriting} tests whether replacing
rendered handwriting typefaces with real glyph instances reduces this remaining
handwriting gap.

\subsection{Where Does Synthetic Rendering Fall Short? Handwriting}
\label{sec:handwriting}

The preceding comparison leaves a clear gap to real handwriting supervision. We test
whether handwriting typefaces are sufficient or whether real glyph
variation provides an additional benefit.
The \textbf{handwriting typefaces} variant samples from the full bank of 693 font families
rather than the measured Thai font profile; 24.0\% of pages use a handwriting typeface, half with
per-instance stroke distortion augmentation. The \textbf{real glyph instances} variant
instead redraws Thai characters on the same subset using the glyph bank in
Section~\ref{sec:font-glyph}. Both variants keep the OCR labels, layout annotations, ink
height, and ink color fixed, and the glyph sources are disjoint from the evaluation sets. Their
comparison therefore isolates the source of the strokes.
Figures~\ref{fig:handwriting-pages} and~\ref{fig:handwriting-detail} in
Appendix~\ref{app:handwriting-examples} show the two variants on the same pages.

\begin{table}[h!]
  \centering
  \caption{Comparison of handwriting rendering. Values are CER in percent (lower is
  better); Med.\ is the page median and Mean is the character-weighted
  CER. The best result within each training block is bold; the last two rows are
  references.}
  \label{tab:handwriting-ink}
  \small
  \setlength{\tabcolsep}{5pt}
  \begin{tabular}{llcccccc}
    \toprule
    & & \multicolumn{2}{c}{Heldout} & \multicolumn{2}{c}{Handwriting}
      & \multicolumn{2}{c}{Easy Handwriting} \\
    \cmidrule(lr){3-4} \cmidrule(lr){5-6} \cmidrule(lr){7-8}
    Training data & Regime & Med. & Mean & Med. & Mean & Med. & Mean \\
    \midrule
    Out-of-Domain Synthetic & page & \textbf{5.07} & \textbf{16.90} & 43.99 & 46.71
      & 38.90 & 44.21 \\
    $+$ handwriting typefaces   & page & 5.89 & 18.44 & 38.65 & 42.83 & 34.84 & 38.11 \\
    $+$ real glyph instances    & page & 6.41 & 20.50 & \textbf{37.91} & \textbf{41.05}
      & \textbf{30.66} & \textbf{37.31} \\
    \midrule
    Out-of-Domain Synthetic & crop & 5.52 & 19.09 & 49.15 & 49.37 & 48.26 & 49.50 \\
    $+$ handwriting typefaces   & crop & 4.51 & 17.68 & 42.44 & 42.54 & 40.64 & 42.15 \\
    $+$ real glyph instances    & crop & \textbf{4.34} & \textbf{16.78} & \textbf{39.97}
      & \textbf{40.25} & \textbf{35.66} & \textbf{37.79} \\
    \midrule
    Real Thai (Print + Handwriting) & page & 1.40 & 8.49 & 26.05 & 28.26 & 20.79 & 25.55 \\
    Qwen3-VL-2B-Instruct    & page & 14.47 & 28.62 & 61.59 & 61.25 & 59.29 & 61.32 \\
    \bottomrule
  \end{tabular}
\end{table}

Handwriting typefaces improve both handwriting sets. Median CER falls from 43.99 to 38.65
on Handwriting and from 38.90 to 34.84 on Easy Handwriting under page-level training. Under
crop-level training, the corresponding changes are 49.15 to 42.44 and 48.26 to 40.64. However, on printed Heldout, the page-level median increases from 5.07 to 5.89, while the crop-level
median decreases from 5.52 to 4.51. Handwriting typefaces therefore improve transfer to
real handwriting in both training settings. On printed Heldout, however, they degrade
page-level performance while improving crop-level performance.

Real glyph instances further improve all handwriting medians, reaching 37.91 and 30.66
under page-level training and 39.97 and 35.66 under crop-level training. Because the two
variants differ only in the source of the strokes, these gains show that real glyph
variation matters beyond the choice of handwriting typeface. The gap to real handwriting
supervision remains: Real Thai (Print + Handwriting) reaches 26.05 on Handwriting and 20.79
on Easy Handwriting. The bank contains only 5,953 instances, leaving the effect of
a larger glyph bank for future study.

\section{From Controlled Findings to Wayu-Paxa-OCR-Zero}
\label{sec:wayuocr}

The controlled studies show that font diversity, document structure, and real glyph
variation improve transfer to real Thai documents. We use these findings to build
Wayu-Paxa-OCR-Zero, a crop-level OCR model trained without OCR labels from real Thai
documents.

\subsection{Wayu-Paxa-OCR-Zero}

We extend the synthetic data used in Section~\ref{sec:transfer} in three ways. First, we
retain both renderings of the 7,131 handwriting pages by including the
handwriting-typeface pages and their real-glyph copies. Second, we add two sets of
handwriting-focused pages: 3,000 disjoint DocLayNet-v1.2 pages on which all text regions
are rendered with handwriting glyphs, and 999 synthetic pages with pasted iApp Handwriting Dataset
crops~\citep{iapp2024thaihandwriting}. Third, we add 2,190 filled forms
reconstructed from public CommonForms templates. The added pages and templates use public
English sources whose OCR labels are translated into Thai. No Real Thai document image
from Section~\ref{sec:transfer} is used as a source page. Handwritten content is either
drawn from the glyph bank in Section~\ref{sec:font-glyph} or inserted as iApp Handwriting
Dataset crops.

\begin{table}[h!]
  \centering
  \caption{Synthetic training data for Wayu-Paxa-OCR-Zero. The three subsets contain
  45,723 generated pages;}
  \label{tab:wayuocr-corpus}
  \small
  \begin{tabular}{lrp{0.24\linewidth}p{0.32\linewidth}}
    \toprule
    Subset & Pages & Source documents & Text rendering \\
    \midrule
    Base synthetic & 39,534 & DocLayNet, Crello, PubTabNet & rendered fonts; 7,131
      handwriting pages also appear with real glyph instances \\
    Handwriting & 3,999 & DocLayNet-v1.2 pages disjoint from the base set & 3,000 pages with
      stitched glyph instances; 999 pages with pasted iApp Handwriting Dataset crops \\
    Forms & 2,190 & CommonForms templates with widget counts matched to real Thai forms
      & stitched glyph instances in every field value \\
    \midrule
    Total & 45,723 & & \\
    \bottomrule
  \end{tabular}
\end{table}

Every page is generated by the pipeline in Section~\ref{sec:synthetic-pipeline} from public
English sources. No Thai document page is used as a source image, and no OCR
label from a real Thai document enters training. Region annotations and OCR labels come
directly from reconstruction outputs, source-region annotations, or known template and
field strings.

We train all 0.9B parameters for one epoch using a learning rate of $3\times10^{-5}$, a
cosine schedule, a warmup ratio of 0.03, and an effective batch size of 16. At inference,
PP-DocLayoutV3 supplies regions to the PaddleOCR-VL recognizer. The
reported scores therefore include layout-detection and reading-order errors.

\subsection{Comparison with Existing Thai OCR Systems}

We compare Wayu-Paxa-OCR-Zero with its base checkpoint, two open Thai OCR systems, and a frontier model to assess the gains from synthetic-only training and its competitiveness with existing systems. We further evaluate on ThaiOCRBench~\citep{nonesung2025thaiocrbench} and SEA-DocBench~\citep{yue2026seavision} to measure generalization beyond our internal evaluation sets.

\begin{table}[h!]
  \centering
  \caption{Comparison with existing Thai OCR systems. Values are CER in percent (lower is
  better); Med.\ is the page median and Mean is the character-weighted CER. ThaiOCRBench
  reports each aggregate averaged across five audited reading tasks; SEA-DocBench is
  evaluated on its 1,148-page clean Thai subset.
  Rules separate the three system families; within each family the better result in each
  column is bold.}
  \label{tab:system-comparison}
  \scriptsize
  \setlength{\tabcolsep}{3pt}
  \begin{tabular}{lcccccccccc}
    \toprule
    & \multicolumn{2}{c}{Heldout} & \multicolumn{2}{c}{Handwriting}
      & \multicolumn{2}{c}{Easy Handwriting} & \multicolumn{2}{c}{ThaiOCRBench}
      & \multicolumn{2}{c}{SEA-DocBench} \\
    \cmidrule(lr){2-3} \cmidrule(lr){4-5} \cmidrule(lr){6-7}
      \cmidrule(lr){8-9} \cmidrule(lr){10-11}
    System & Med. & Mean & Med. & Mean & Med. & Mean & Med. & Mean & Med. & Mean \\
    \midrule
    PaddleOCR-VL-1.6 (0.9B) & 6.64 & 27.60 & 74.87 & 67.90 & 73.74 & 69.41
      & 38.0 & 43.3 & 8.87 & 18.54 \\
    \ourrow Wayu-Paxa-OCR-Zero (0.9B)        & \textbf{1.24} & \textbf{14.75} & \textbf{20.55}
      & \textbf{22.28} & \textbf{14.18} & \textbf{17.08} & \textbf{15.3}
      & \textbf{25.6} & \textbf{4.86} & \textbf{9.99} \\
    \midrule
    Typhoon OCR (7B)                 & 2.54 & 18.27 & 43.60 & 49.36 & 34.99 & 45.06
      & 30.6 & 44.7 & 9.22 & 17.57 \\
    Typhoon OCR 1.5 (2B)             & \textbf{0.21} & \textbf{5.47} & \textbf{19.36}
      & \textbf{21.86} & \textbf{9.02} & \textbf{15.74} & \textbf{6.2}
      & \textbf{16.8} & \textbf{5.81} & \textbf{12.80} \\
    \midrule
    Gemini 3.7 Flash                 & 0.00 & 3.56 & 11.29 & 15.04 & 3.89 & 7.64
      & 0.9 & 3.8 & 5.51 & 10.62 \\
    \bottomrule
  \end{tabular}
\end{table}

Comparison with the base checkpoint isolates the effect of our training recipe because the
architecture is fixed. As Table~\ref{tab:system-comparison} shows, Wayu-Paxa-OCR-Zero reduces
both median and mean CER on all five benchmarks. On the internal sets, median CER falls from
6.64\% to 1.24\% on Heldout, from 74.87\% to 20.55\% on Handwriting, and from 73.74\% to
14.18\% on Easy Handwriting. The handwriting reductions show that synthetic supervision
substantially improves recognition on a typical page, rather than only correcting a small
number of failures. Mean CER nevertheless remains above median CER on every benchmark, which
is consistent with harder examples contributing disproportionately to aggregate error.

Comparisons with independently trained systems instead measure competitiveness, because the
systems differ in scale, pretraining, and supervision. Despite using 0.9B parameters,
Wayu-Paxa-OCR-Zero outperforms the 7B Typhoon OCR model on all five benchmarks. It nearly
matches the 2B Typhoon OCR 1.5 model on Handwriting (20.55\% versus 19.36\% median CER) and
performs better on SEA-DocBench (4.86\% versus 5.81\%), but remains behind on Heldout, Easy
Handwriting, and ThaiOCRBench. Gemini 3.7 Flash remains strongest on the three internal sets
and ThaiOCRBench, whereas Wayu-Paxa-OCR-Zero achieves lower CER on SEA-DocBench. These
rankings establish competitiveness; they do not show that synthetic data are intrinsically
superior to the supervision used by the other systems.

This pattern is consistent with the controlled experiments in Section~\ref{sec:transfer}:
synthetic reconstruction approaches real printed supervision under matched conditions but
does not fully close the gap, and transfer depends on typeface diversity, two-dimensional
structure, glyph variation, and training granularity. The remaining differences across
benchmarks therefore indicate that synthetic-only training is already competitive, while
broader coverage of document layouts, typography, and handwriting variation remains an
important direction for improving generalization.

\section{Discussion, Limitations, and Conclusion}
\label{sec:discussion}

Our results show that synthetic reconstruction can produce competitive Thai OCR without
page-level OCR labels from real Thai documents. Typeface diversity, two-dimensional structure,
and real handwriting glyphs improve transfer, whereas non-text page context has little
consistent effect. However, synthetic supervision still trails real supervision on severe
failures and handwriting, and the value of in-domain reconstruction depends on training
granularity.

We hope this work encourages further research on synthetic OCR for Thai and other languages
with limited document annotations.

\section*{Acknowledgments}

This work is a collaboration between Wayu Research and Paxa Labs and was self-funded by Wayu Research. We would also like to thank Surapon Nonsung and the Typhoon Team for their valuable feedback on this technical report, as well as the global and local AI communities for open-sourcing resources and sharing knowledge that made this work possible.

\section*{Ethics Statement}

We use publicly available Thai documents, fonts, and handwriting samples to study synthetic
document reconstruction. Replacing original text with generated content can reduce exposure
of personal information, but models trained on synthetic data may be less robust than those
trained on real data, especially for unseen document types, typefaces, and handwriting
styles. Released artifacts should document their sources, intended uses, limitations, and
privacy safeguards.

\bibliography{colm2026_conference}
\bibliographystyle{colm2026_conference}

\appendix
\section{Training and Inference Formats}
\label{app:formats}

\paragraph{Page-level format.}
Page-level training uses one example per document page. The input contains the page image
and a fixed instruction, and the target is a single JSON object. The instruction is

\begin{quote}\small\ttfamily
Extract this document page. Return a JSON object with one field: "layout" --- a list of
blocks in natural reading order, each \{"bbox":[x1,y1,x2,y2], "category":<Title$|$
Section-header$|$Text$|$List-item$|$Table$|$Formula$|$Picture$|$Caption$|$Footnote$|$
Page-header$|$Page-footer>, "text":<Markdown; HTML for a Table; LaTeX for a Formula>\}.
\end{quote}

\noindent The corresponding target has the following form:

\begin{quote}\small\ttfamily
\{"layout":[\{"bbox":[71,54,929,103],"category":"Page-header","text":$\ldots$\},
\{"bbox":[71,142,929,388],"category":"Text","text":$\ldots$\}]\}
\end{quote}

\noindent Box coordinates are normalized to $[0,1000]$ on both axes, and the eleven
categories follow the DocLayNet label set. \texttt{Picture} blocks contain no text,
\texttt{Table} blocks use HTML, \texttt{Formula} blocks use LaTeX, and all other blocks use
Markdown. A single decode therefore predicts the box geometry, category, reading order,
and text for the complete page; malformed output affects the complete page prediction.

\paragraph{Crop-level format.}
Crop-level training uses one example per document region. The input is a crop from the
annotated box, and the target is the OCR label of that region. We add a 1\% margin, encode
the crop as JPEG to match the inference input, and remove crops whose width or height is
below 8 pixels. We exclude \texttt{Picture} regions. Each category uses the prompt assigned
by PaddleX: \texttt{Table Recognition:} for \texttt{Table}, \texttt{Formula Recognition:}
for \texttt{Formula}, \texttt{Chart Recognition:} for \texttt{Chart}, and \texttt{OCR:}
otherwise. \texttt{Formula} crops also use the PaddleX margin trim.

The Qwen3-VL and PaddleOCR-VL crop models differ only in their table targets. Qwen3-VL uses
the HTML targets from the page-level format, whereas PaddleOCR-VL converts them to
Optimized Table Structure Language (OTSL)~\citep{lysak2023optimized}, the serialization
used during its pretraining. We remove tables whose spans cannot be represented faithfully
in OTSL. At inference, the layout detector supplies the regions, so the reported crop
scores include detection and reading-order errors and measure the complete pipeline.

\section{Real Thai OCR Label Construction}
\label{app:real-labels}

We construct pseudo-labels for Real Thai (Print) and Real Thai (Handwriting) using a
two-stage pipeline inspired by Typhoon OCR~\citep{nonesung2026typhoonocr}. Azure OCR first
extracts an initial pseudo-label from each page. A VLM then reads the page together with the
Azure output and normalizes it into the OCR label format used in this study.

\begin{itemize}
  \item \textbf{Training labels:} For the training splits of Real Thai (Print) and Real Thai
    (Handwriting), we use the open-source VLM \texttt{gemma-4-31b-it} to normalize the Azure
    OCR pseudo-labels. These labels provide the real printed and handwriting supervision in
    Section~\ref{sec:real-supervision} and the Thai OCR labels used for In-Domain Synthetic
    in Section~\ref{sec:in-domain}.
  \item \textbf{Evaluation labels:} For Heldout, Handwriting, and Easy Handwriting, we use
    Gemini 2.5 Flash to normalize the Azure OCR pseudo-labels. The resulting labels are the
    references used for all reported CER values.
\end{itemize}

\section{Projected CER}
\label{app:projected-cer}

Page-level systems emit the complete page, whereas crop-level systems emit only the regions
their layout detector returns. Scoring both against one reference therefore charges the
output contract as recognition error: page furniture and undetected blocks are counted as
insertions for one setting and as deletions for the other. Algorithm~\ref{alg:projected-cer}
removes this term by projecting each prediction onto the evaluated regions before scoring.
Every reported CER value uses this procedure.

The prediction $p$ is one string per page: the predicted region texts in predicted reading
order for page-level systems, and the assembled pipeline output for crop-level systems. The
regions $R$ are the reference regions of the evaluated view, in reading order, and the
reference $g$ is their concatenation; the view keeps the text-bearing regions the layout
detector returns and excludes non-text regions and running furniture. Pages whose view is
empty are excluded. Text outside the claimed spans is discarded, while
misreads and hallucinations inside a claimed span, and regions the prediction never
produced, are still charged.

\begin{algorithm}[h]
\caption{Projected CER. $\mathrm{Lev}$ is Levenshtein distance; $\mathrm{Ratio}$ and
$\mathrm{Align}$ are the \texttt{rapidfuzz} indel ratio and its best-matching-substring
alignment.}
\label{alg:projected-cer}
\begin{algorithmic}[1]
\Function{Normalize}{$s$}
  \State remove \texttt{<figure>}\ldots\texttt{</figure>}, \texttt{<page\_number>}\ldots\texttt{</page\_number>}, remaining HTML tags, and \texttt{\$}\ldots\texttt{\$} formulas
  \State remove Markdown and checkbox characters, then all whitespace
  \State \Return NFC-normalized $s$
\EndFunction
\Statex
\Function{Project}{$p$, $R=(r_1,\ldots,r_n)$}
  \State $p \gets \Call{Normalize}{p}$;\quad $t_i \gets \Call{Normalize}{r_i}$, dropping empty $t_i$
  \State $F \gets \{(1,|p|)\}$;\quad $C \gets \emptyset$ \Comment{unclaimed spans; claims}
  \For{$i$ in indices sorted by decreasing $|t_i|$} \Comment{long regions pin their span first}
    \State $\mathit{best} \gets \textbf{nil}$
    \For{$(s,e) \in F$ with $w \gets p[s..e] \neq \varepsilon$}
      \If{$|w| \leq |t_i|$}
        $(\sigma,c) \gets (\Call{Ratio}{t_i,w},\,(s,e))$ \Comment{no room; offer the gap}
      \Else{}
        $(\sigma,c) \gets \Call{Align}{t_i,w}$ \Comment{score and span of the best substring}
      \EndIf
      \If{$c \neq \varepsilon$ and ($\mathit{best} = \textbf{nil}$ or $\sigma > \sigma_{\mathit{best}}$)}
        $\mathit{best} \gets (\sigma, c, (s,e))$
      \EndIf
    \EndFor
    \If{$\mathit{best} = \textbf{nil}$} \textbf{continue} \EndIf
    \State $(\sigma,(c_s,c_e),(f_s,f_e)) \gets \mathit{best}$
    \State $C \gets C \cup \{(i,c_s,c_e)\}$ \Comment{a claimed span cannot be claimed again}
    \State $F \gets \big(F \setminus \{(f_s,f_e)\}\big) \cup \{(f_s,c_s),(c_e,f_e)\}$, keeping nonempty spans
  \EndFor
  \State \Return concatenation of $p[c_s..c_e]$ over $C$, $i$ increasing \Comment{reading order}
\EndFunction
\Statex
\Function{PageCER}{$p$, $R$, $g$}
  \State $g \gets \Call{Normalize}{g}$;\quad $L \gets |g|$
  \State \Return $\min\!\big(\mathrm{Lev}(\Call{Project}{p,R},\,g) \,/\, \max(L,1),\; 1\big)$, $\;L$
\EndFunction
\Statex
\State $(\mathit{cer}_j, L_j) \gets \Call{PageCER}{p_j,R_j,g_j}$ for every page $j$ of the evaluation set
\State \textbf{Median} $\gets \mathrm{median}_j\, \mathit{cer}_j$;\quad
       \textbf{Mean} $\gets \sum_j \mathit{cer}_j L_j \,/\, \sum_j L_j$
\end{algorithmic}
\end{algorithm}

\section{Handwriting Glyph Bank Construction}
\label{app:glyph-bank}

We construct a single handwriting glyph bank from line images drawn from the iApp
Handwriting Dataset~\citep{iapp2024thaihandwriting} and the Real Thai (Handwriting) training
split. We exclude the handwriting evaluation split. The construction pipeline has four
stages:
\begin{itemize}
  \item \textbf{Glyph detection:} We train a class-agnostic detector based on
    \href{https://docs.ultralytics.com/models/yolov8/}{YOLOv8} and apply it to each line
    crop. For every candidate, we retain a tight ink mask, its vertical band (above, main,
    or below), and the line text height used for scale normalization. This stage produces
    5.70M candidates.
  \item \textbf{OCR-label alignment and filtering:} We align the candidates monotonically
    with the shaped line OCR label while accounting for their vertical bands. The alignment
    removes duplicate, spurious, and merged boxes and maps each remaining candidate to a
    character. We retain a mapped identity when it is confirmed by a YOLOv8-based glyph
    classifier, a commercial OCR reading at the same position, or both. Rare classes
    require OCR confirmation because the classifier is less reliable for these classes.
    Size thresholds and class-specific shape filters remove remaining fragments, leaving
    534,404 candidates.
  \item \textbf{Vision--language verification:} We sample the retained candidates and
    classify them with Gemini 3.5 Flash. The model confirms 12,090 instances.
  \item \textbf{Bank assembly:} We group the instances by character class and
    select at most 60 instances per class by round-robin sampling across writers. The
    resulting bank contains 5,953 instances across 76 character classes.
\end{itemize}

\begin{table}[h!]
  \centering
  \caption{Handwriting glyph-bank construction. Alignment and filtering combine OCR-label
  alignment, classifier or OCR confirmation, and size and shape filtering. Vision--language
  verification retains crops whose predicted character matches the mapped identity. Each
  class contains at most 60 instances.}
  \label{tab:glyph-funnel}
  \small
  \begin{tabular}{llr}
    \toprule
    Stage & Quantity & Total \\
    \midrule
    Glyph detection         & Candidate boxes      & 5,698,871 \\
    \addlinespace[2pt]
    Alignment and filtering & Retained candidates & 534,404 \\
    \addlinespace[2pt]
    VLM verification        & Verified instances   & 12,090 \\
    \addlinespace[2pt]
    Bank assembly           & Bank instances       & 5,953 \\
                            & Character classes    & 76 \\
    \bottomrule
  \end{tabular}
\end{table}

During real-glyph rendering, we sample each supported character occurrence independently
from the bank and stitch the glyph into the target region at the selected ink height and
color. Sampling does not enforce writer consistency within a page. Characters absent from
the bank use the page typeface. Across the redrawn pages in Section~\ref{sec:handwriting},
the bank covers 96.3\% of Thai character instances.

\section{Source-Property Ablation Examples}
\label{app:ablation-examples}

\begin{figure}[h!]
  \centering
  \setlength{\tabcolsep}{1.2pt}
  \setlength{\fboxsep}{0pt}
  \newcommand{\ablpanel}[1]{%
    \fcolorbox{black!35}{white}{\includegraphics[width=0.186\textwidth]{figures/#1}}}
  \newcommand{\ablhead}[2]{%
    \begin{tabular}{@{}c@{}}\scriptsize #1\\[-2pt]\tiny\itshape #2\end{tabular}}
  \begin{tabular}{@{}ccccc@{}}
    \ablhead{Source document}{DocLayNet, English} &
    \ablhead{Out-of-Domain Synth.}{Thai reconstruction} &
    \ablhead{$-$ non-text context}{White-layout} &
    \ablhead{$-$ font diversity}{White-single-font} &
    \ablhead{$-$ 2-D layout}{Linear-white-single-font} \\[2pt]
    \ablpanel{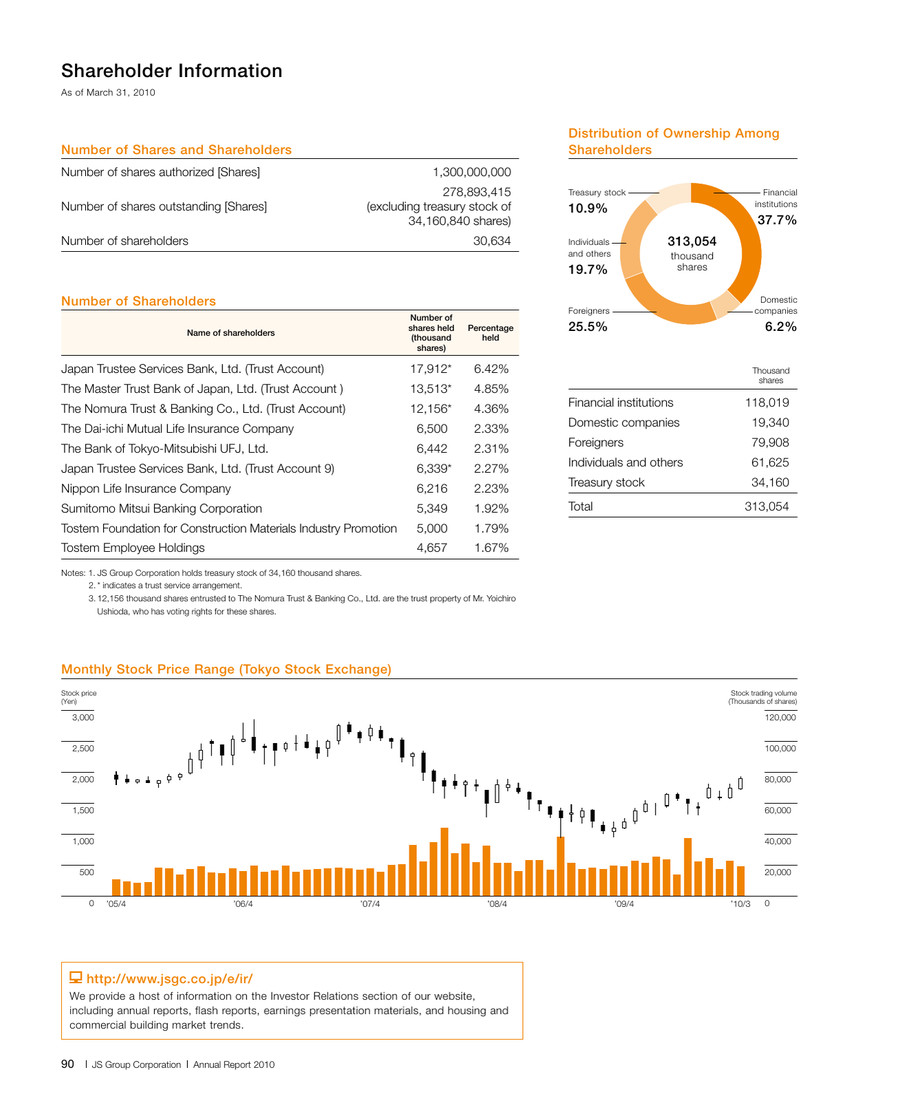} & \ablpanel{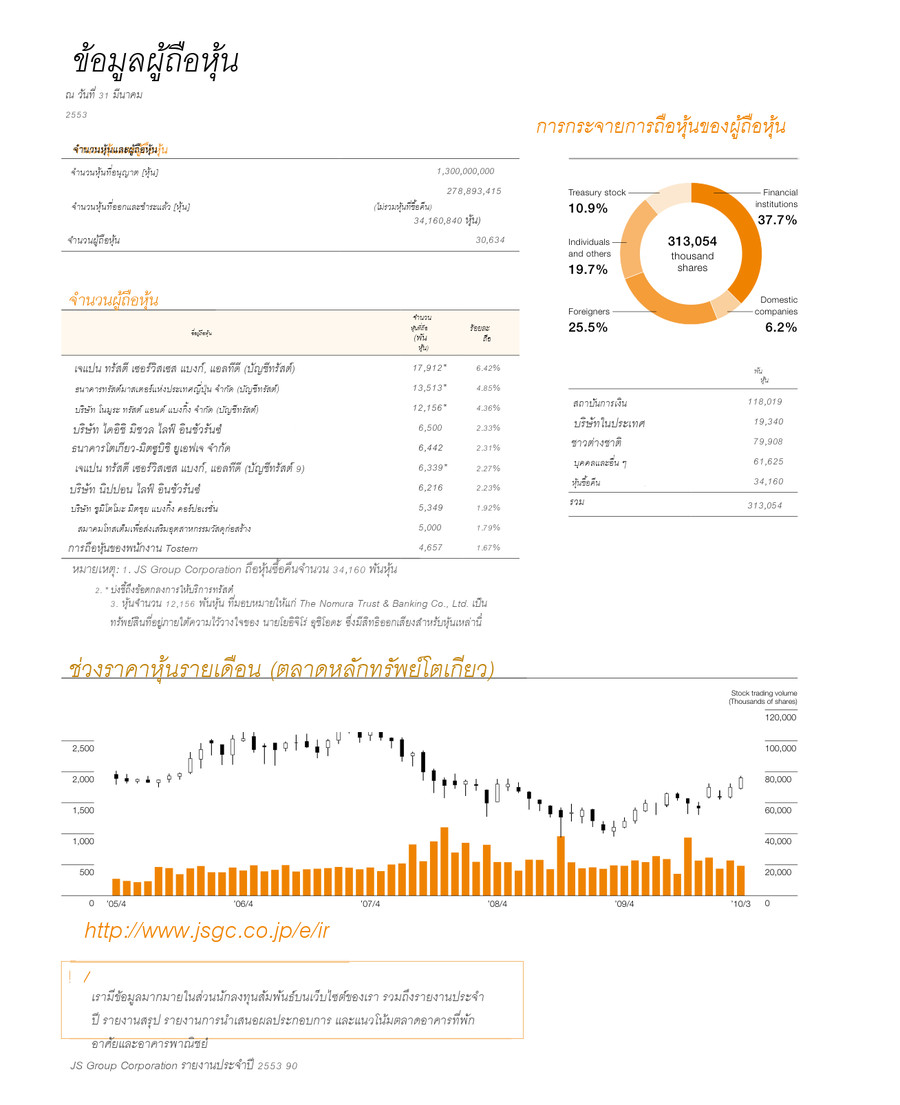} & \ablpanel{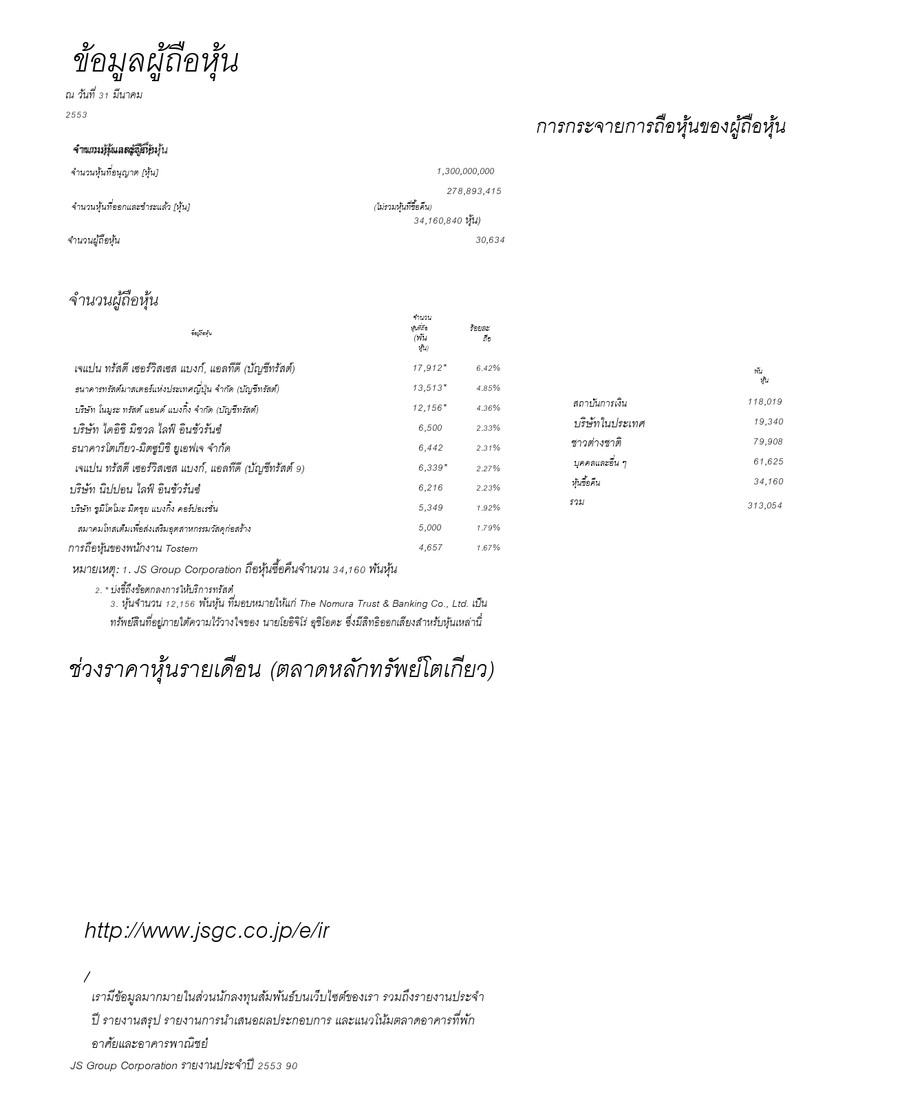} &
    \ablpanel{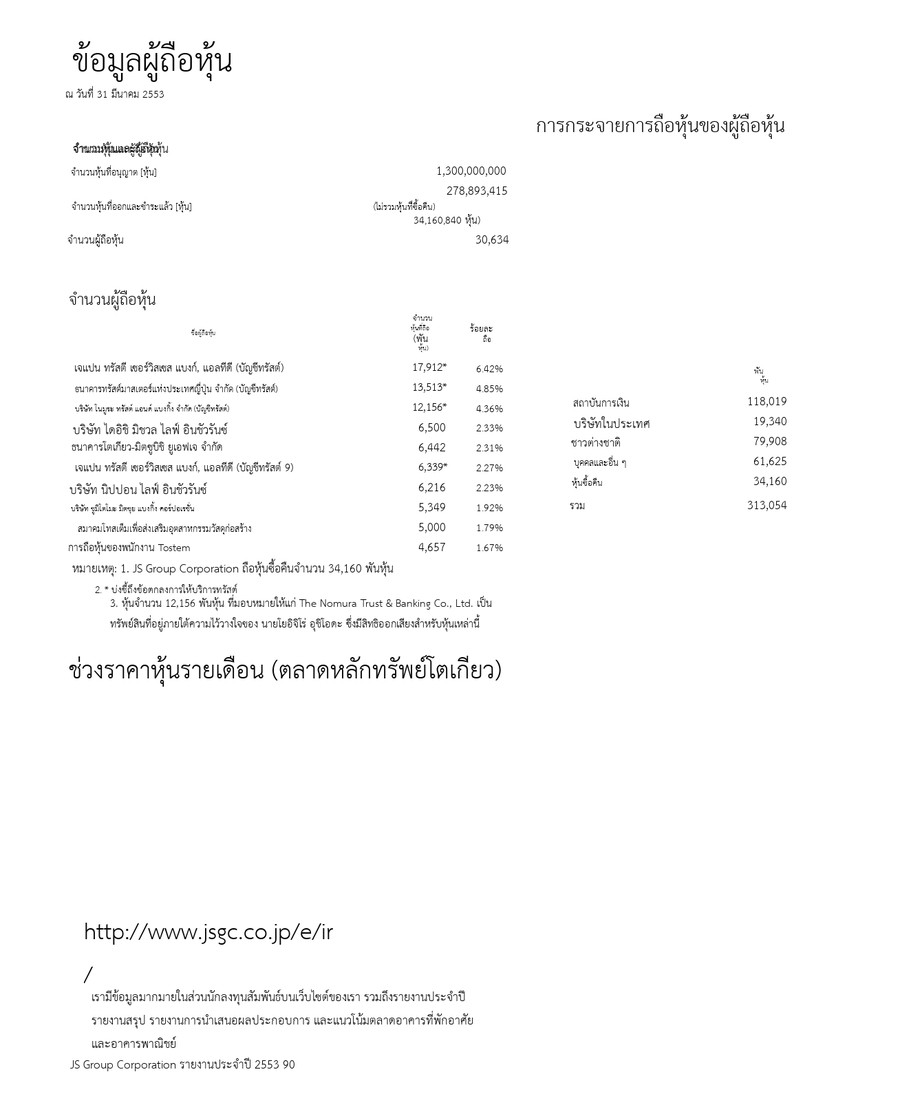} & \ablpanel{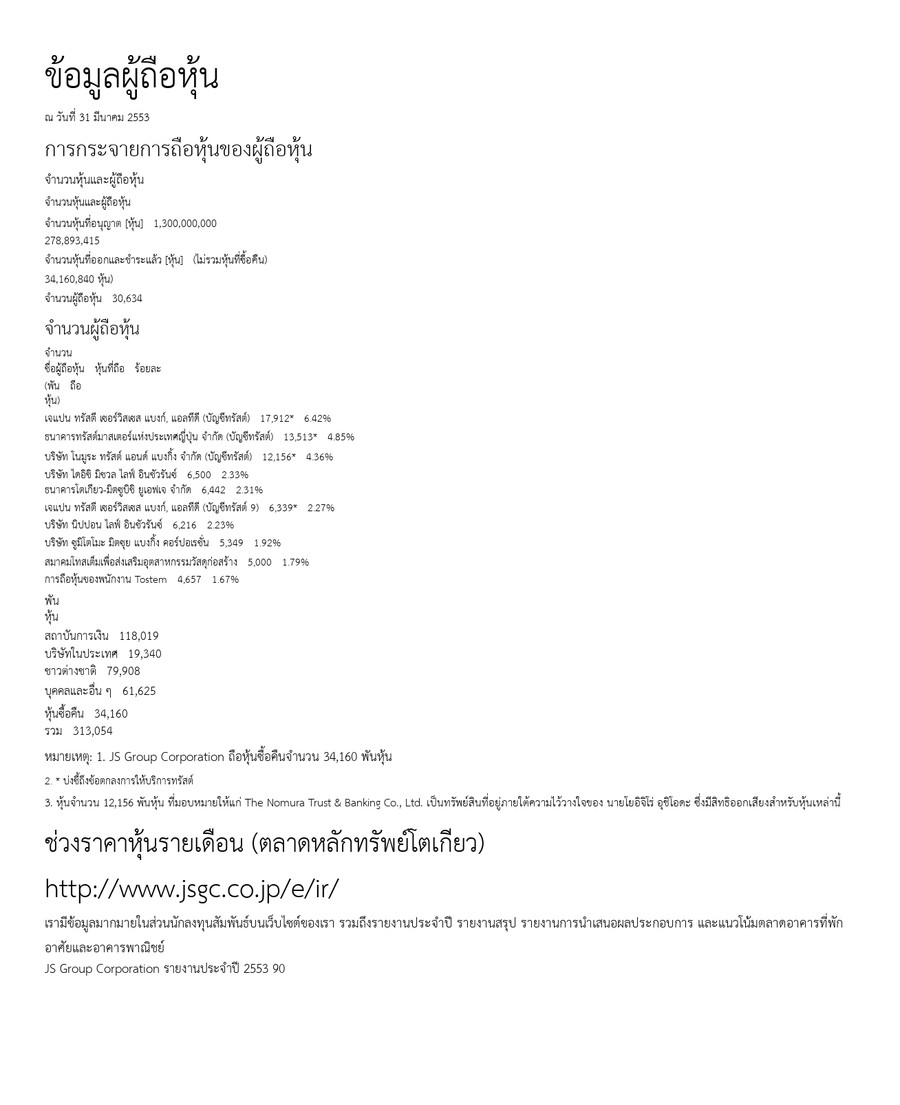} \\[3pt]
    \ablpanel{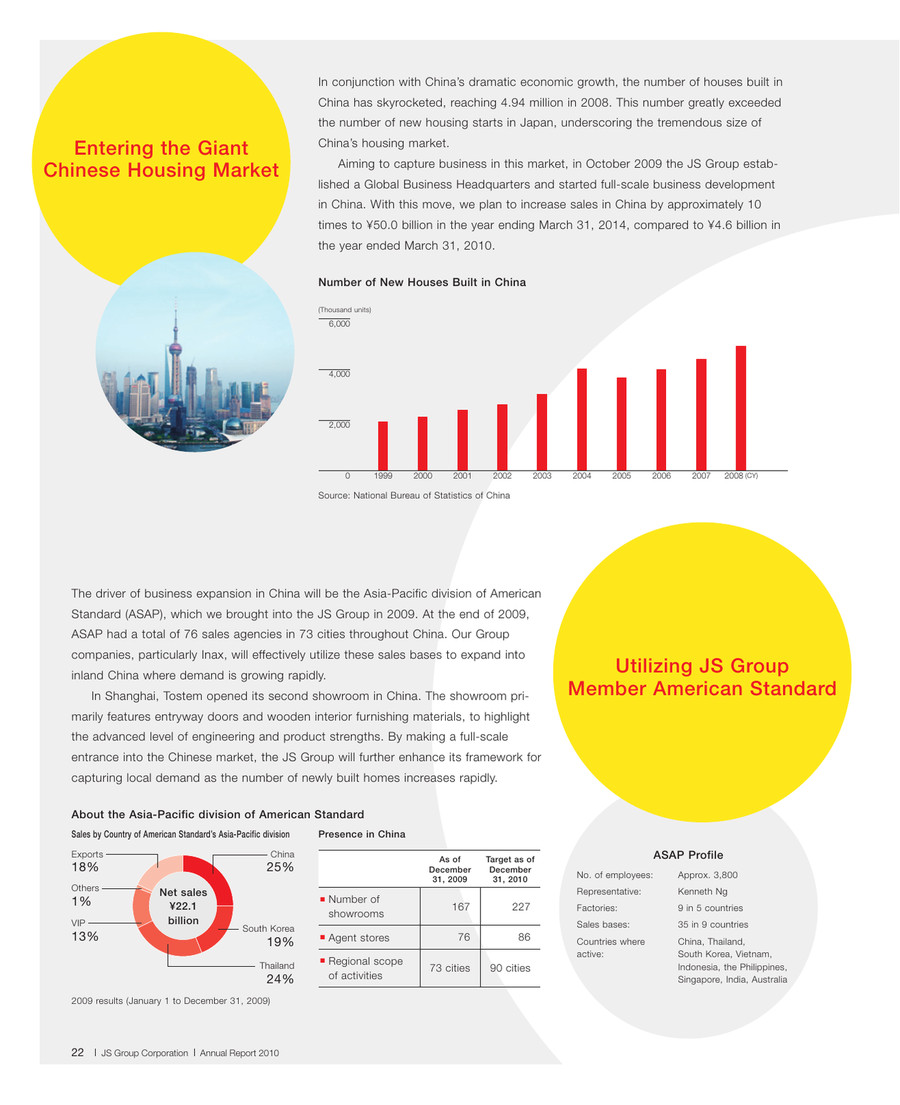} & \ablpanel{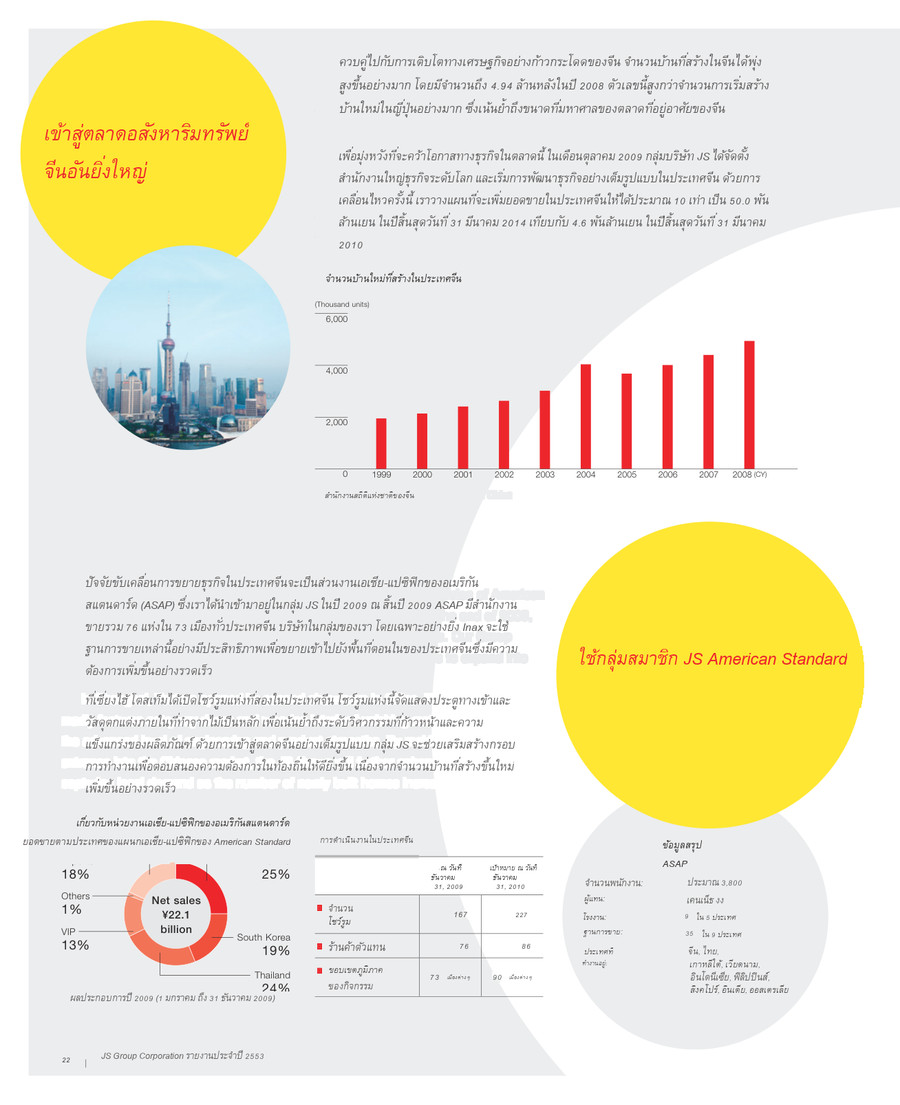} & \ablpanel{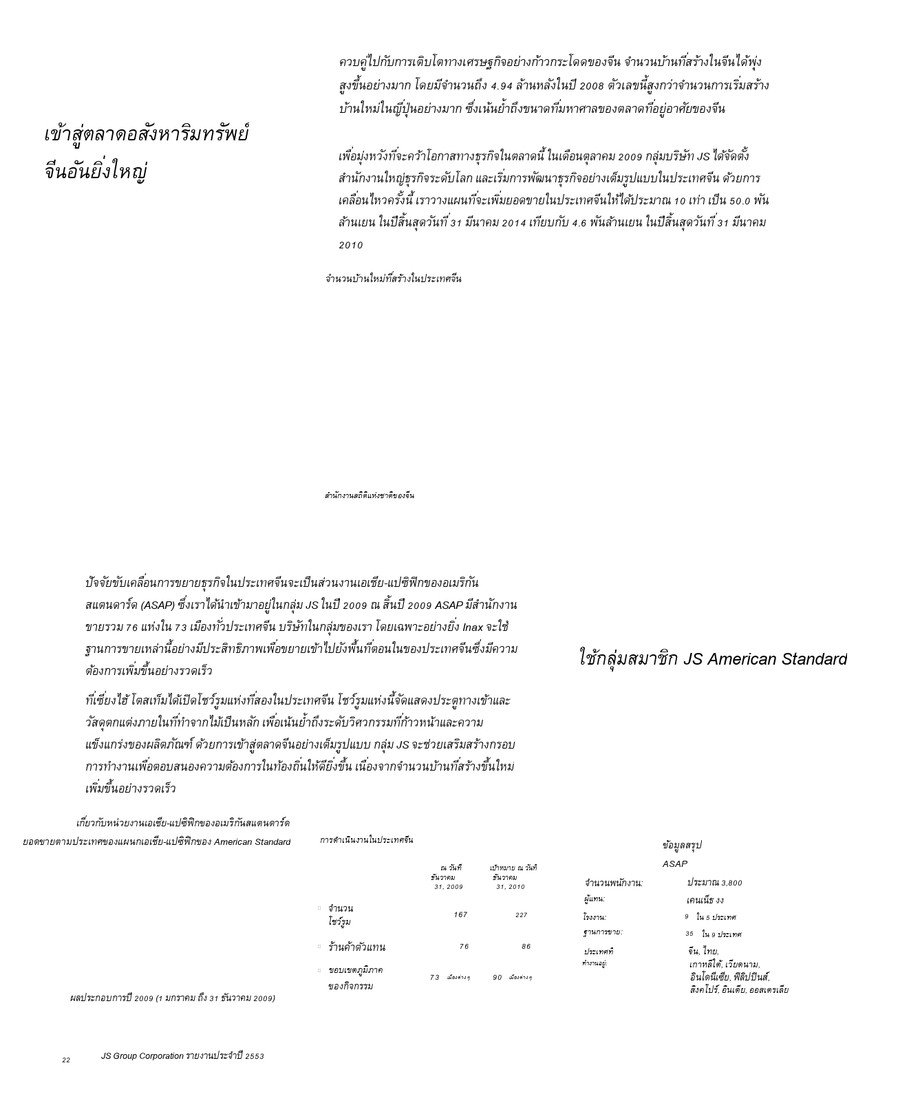} &
    \ablpanel{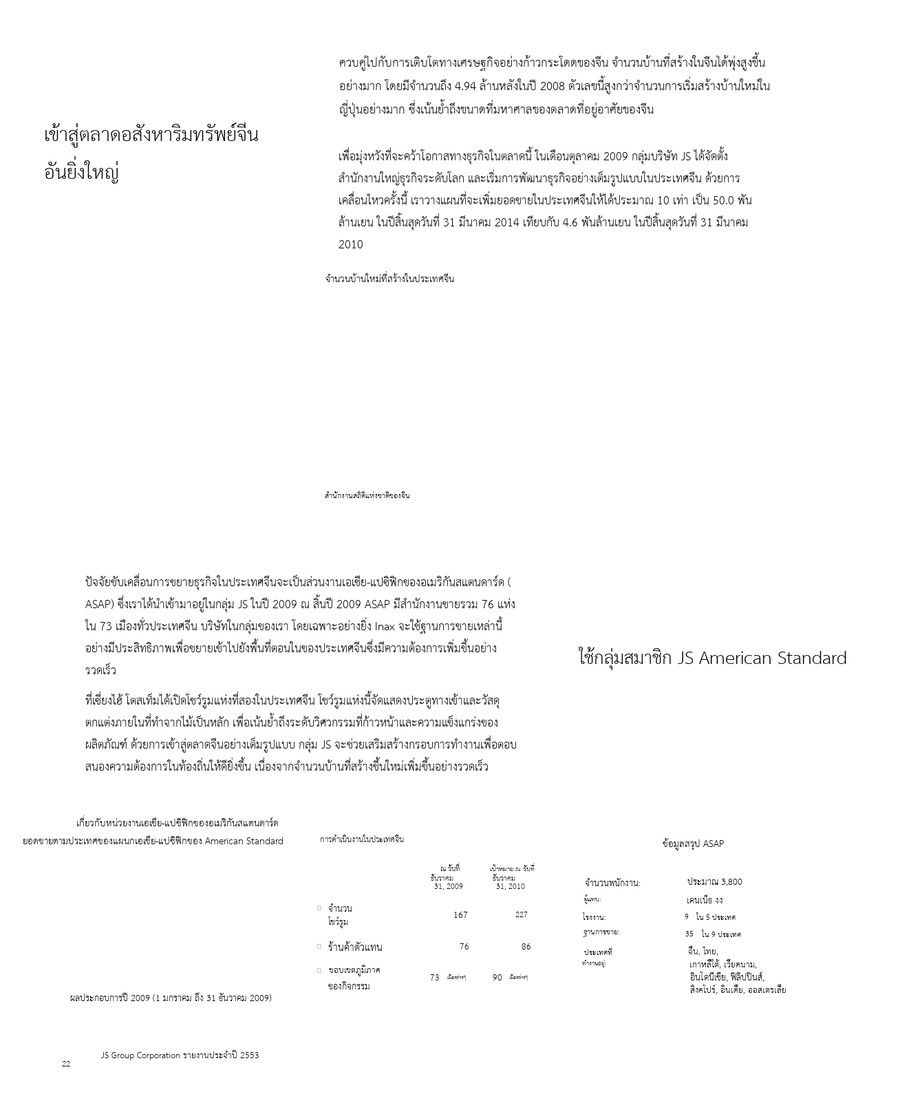} & \ablpanel{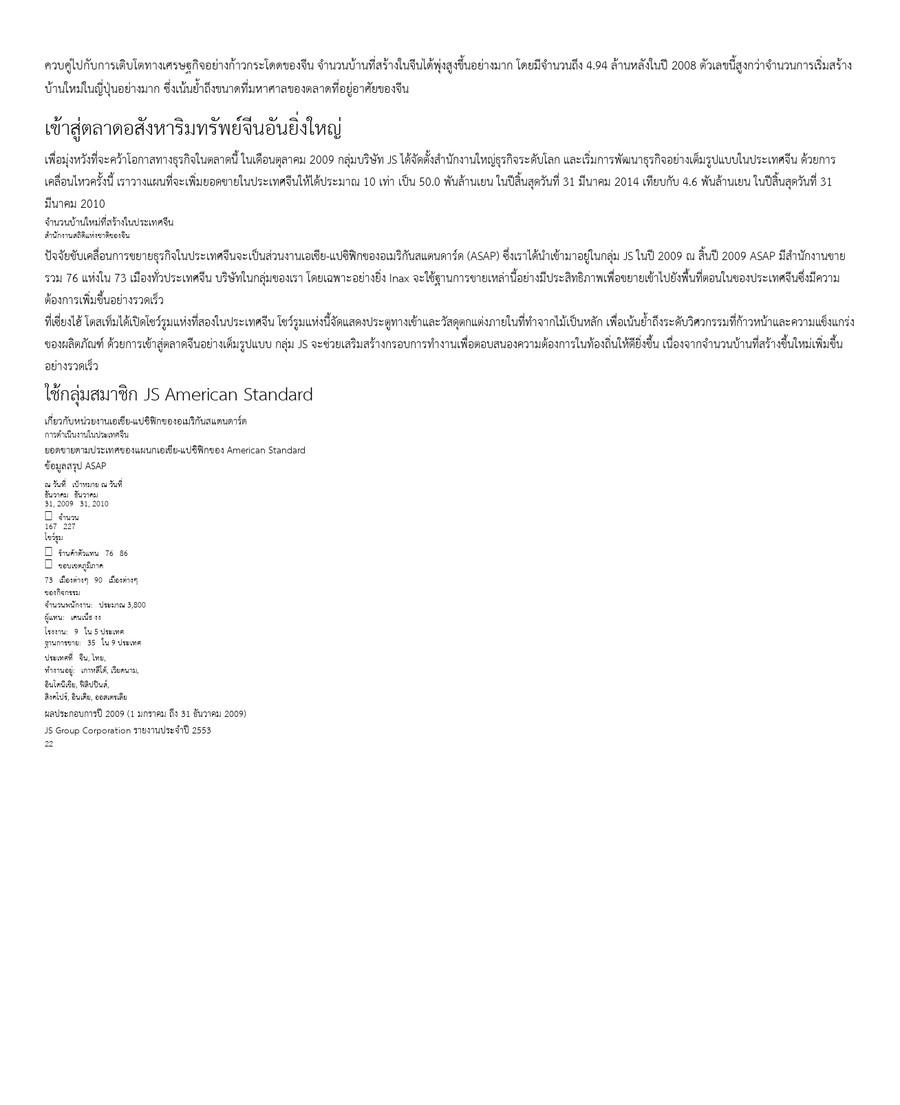} \\
  \end{tabular}
  \caption{Two pages under the source-property variants of
  Section~\ref{sec:source-factors}, ordered from left to right. Out-of-Domain Synthetic
  replaces the English text with Thai while preserving the source layout and non-text
  pixels. \textbf{White-layout} removes the non-text context while retaining every text
  region. \textbf{White-single-font} additionally renders every line in TH Sarabun New
  while preserving its ink height. \textbf{Linear-white-single-font} stacks the regions
  vertically in reading order, removing the two-dimensional layout. Both source pages are
  financial reports from DocLayNet.}
  \label{fig:ablation-examples}
\end{figure}

\begin{figure}[h!]
  \centering
  \setlength{\fboxsep}{0pt}
  \newcommand{\detpanel}[1]{%
    \fcolorbox{black!35}{white}{\includegraphics[width=0.97\textwidth]{figures/#1}}}
  \begin{tabular}{@{}l@{}}
    \scriptsize Source document \textit{(DocLayNet, English)} \\[1pt]
    \detpanel{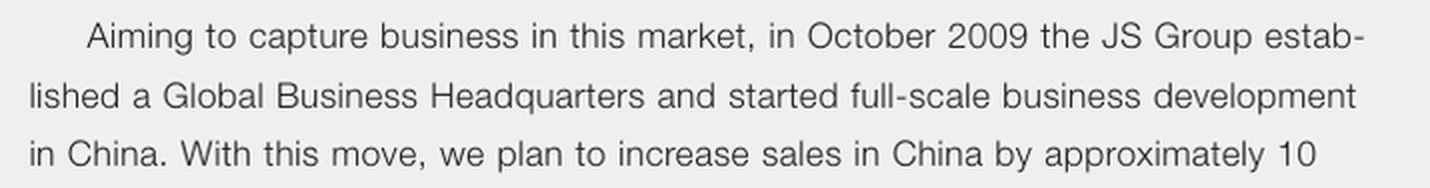} \\[4pt]
    \scriptsize Out-of-Domain Synthetic \textit{(Thai reconstruction)} \\[1pt]
    \detpanel{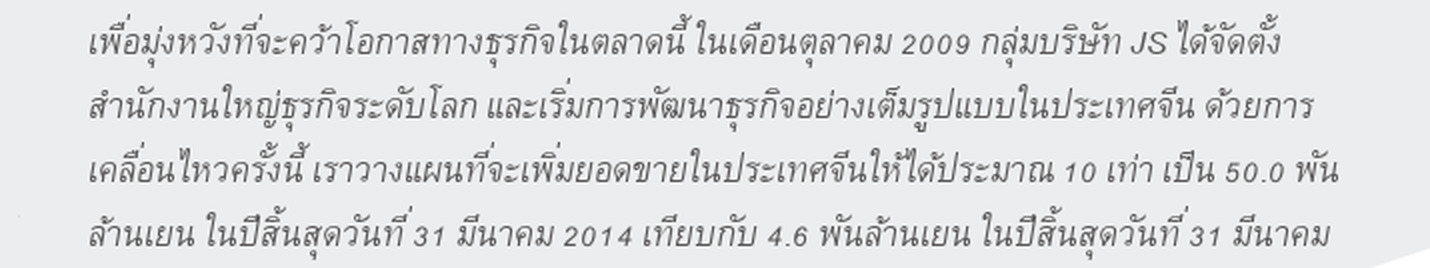} \\[4pt]
    \scriptsize $-$ non-text context \textit{(White-layout)} \\[1pt]
    \detpanel{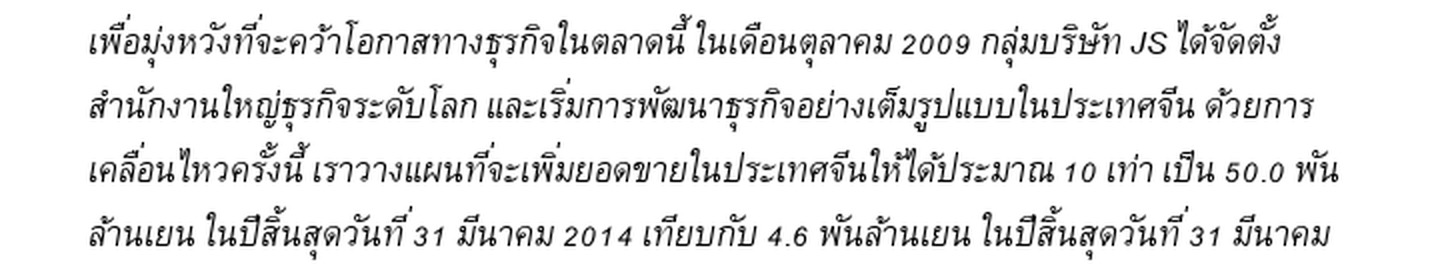} \\[4pt]
    \scriptsize $-$ font diversity \textit{(White-single-font)} \\[1pt]
    \detpanel{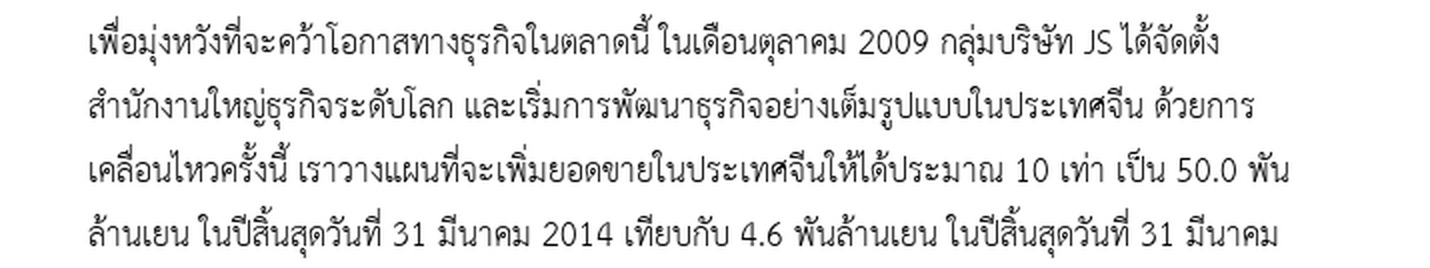} \\
  \end{tabular}
  \caption{The same text region of the second page of Figure~\ref{fig:ablation-examples},
  shown at the same magnification. Out-of-Domain Synthetic replaces the English paragraph
  with its Thai OCR label using a typeface and size selected by the renderer. Removing
  non-text context changes only the region background. Removing font diversity replaces the
  sampled typeface with TH Sarabun New while preserving ink height. The final variant is
  omitted because it changes region placement rather than text appearance.}
  \label{fig:ablation-detail}
\end{figure}

Figures~\ref{fig:ablation-examples} and~\ref{fig:ablation-detail} show the variants of
Section~\ref{sec:source-factors}. Each row follows one English source page through
Out-of-Domain Synthetic and the three successive removals. All four synthetic panels use
the same Thai OCR labels, region boxes, and reading order. We render each variant from
the per-line records of the reconstruction, so adjacent panels differ only in the property
removed. The panels are the images used for training.

\section{Source-Domain Examples}
\label{app:domain-examples}

\begin{figure}[h!]
  \centering
  \setlength{\tabcolsep}{1.5pt}
  \setlength{\fboxsep}{0pt}
  \newcommand{\dompanel}[1]{%
    \fcolorbox{black!35}{white}{\includegraphics[height=2.3in]{figures/#1}}}
  \newcommand{\domhead}[2]{%
    \begin{tabular}{@{}c@{}}\footnotesize #1\\[-2pt]\scriptsize\itshape #2\end{tabular}}
  \begin{tabular}{@{}ccc@{}}
    \domhead{Out-of-Domain Synthetic}{reconstructed English source} &
    \domhead{Real Thai (Print)}{crawled page with original OCR labels} &
    \domhead{In-Domain Synthetic}{reconstructed center page} \\[2pt]
    \dompanel{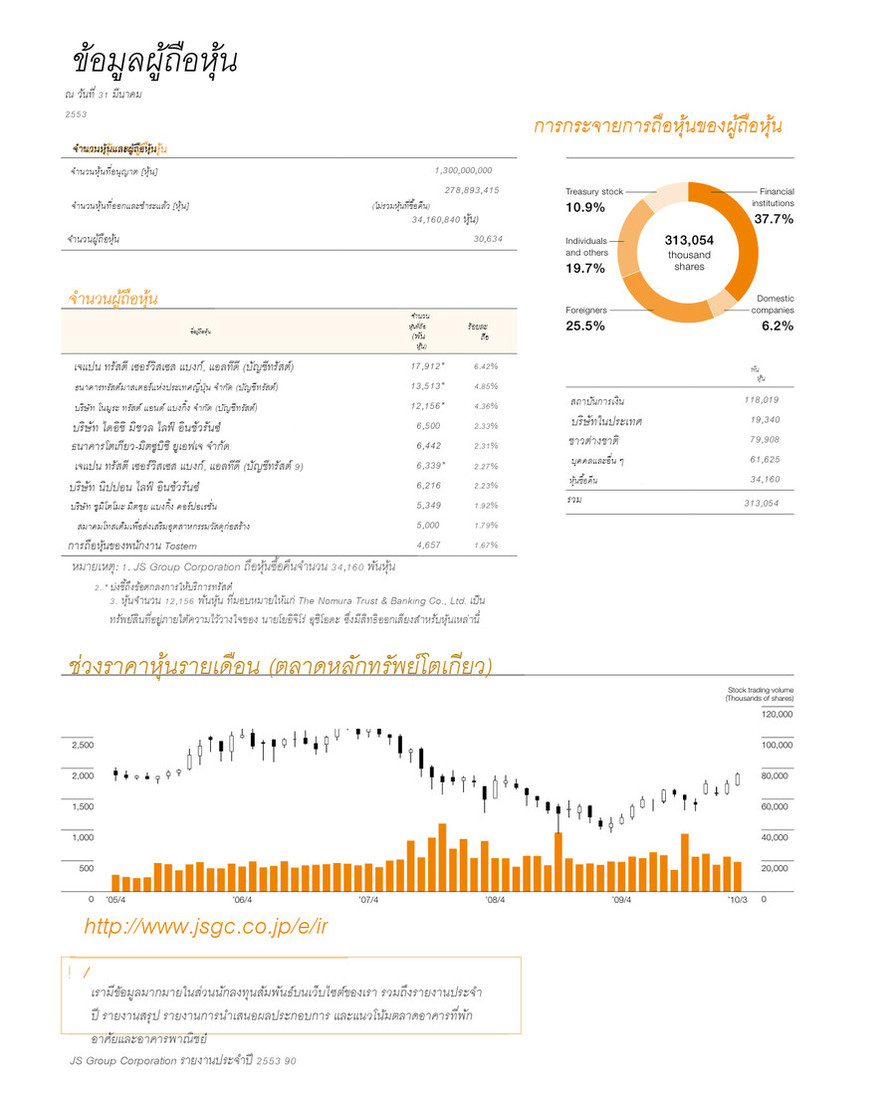} & \dompanel{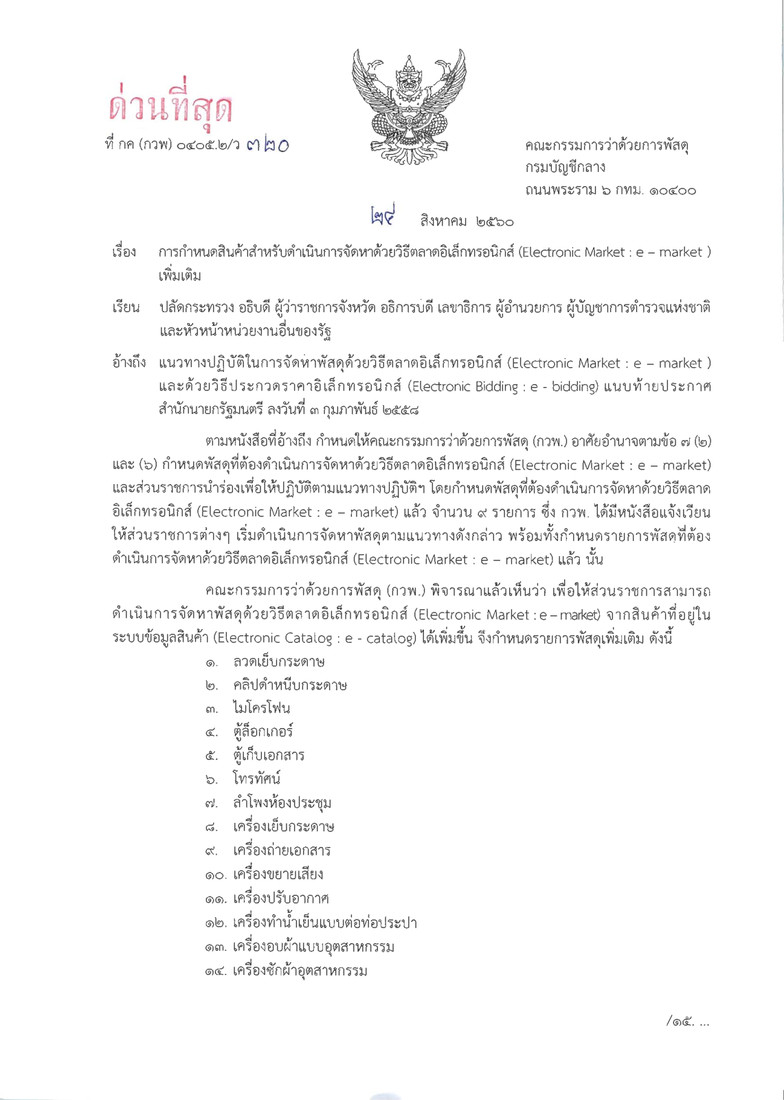} & \dompanel{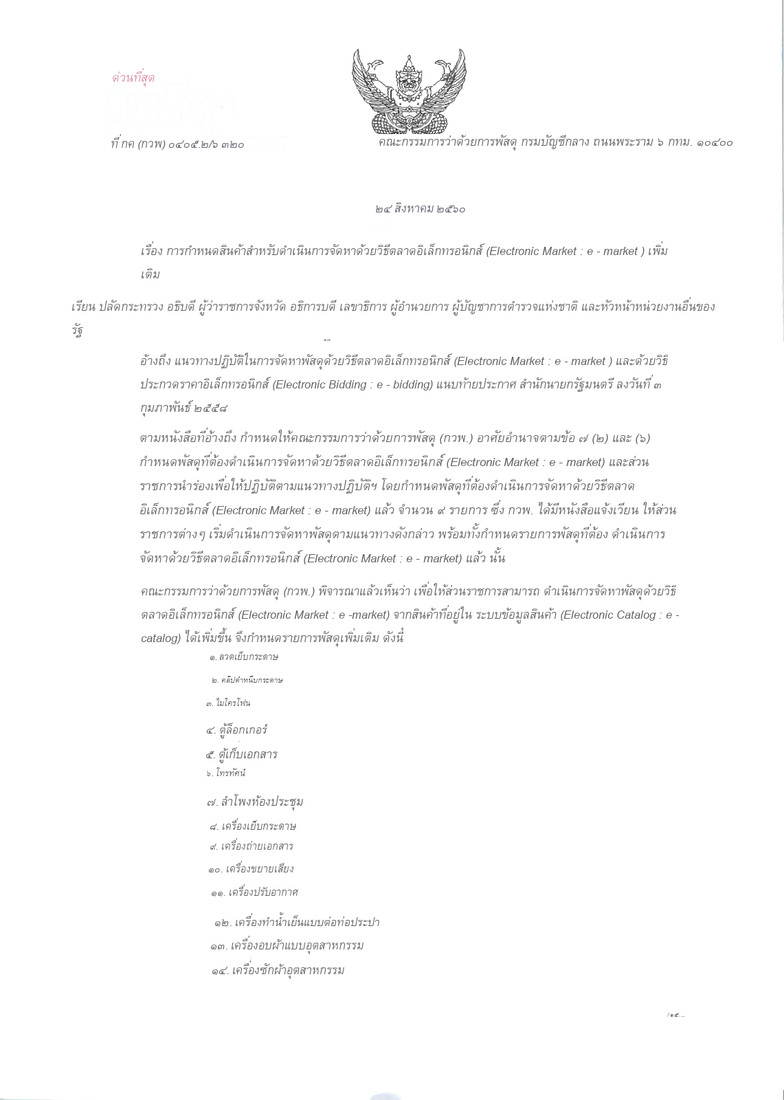} \\[3pt]
    \dompanel{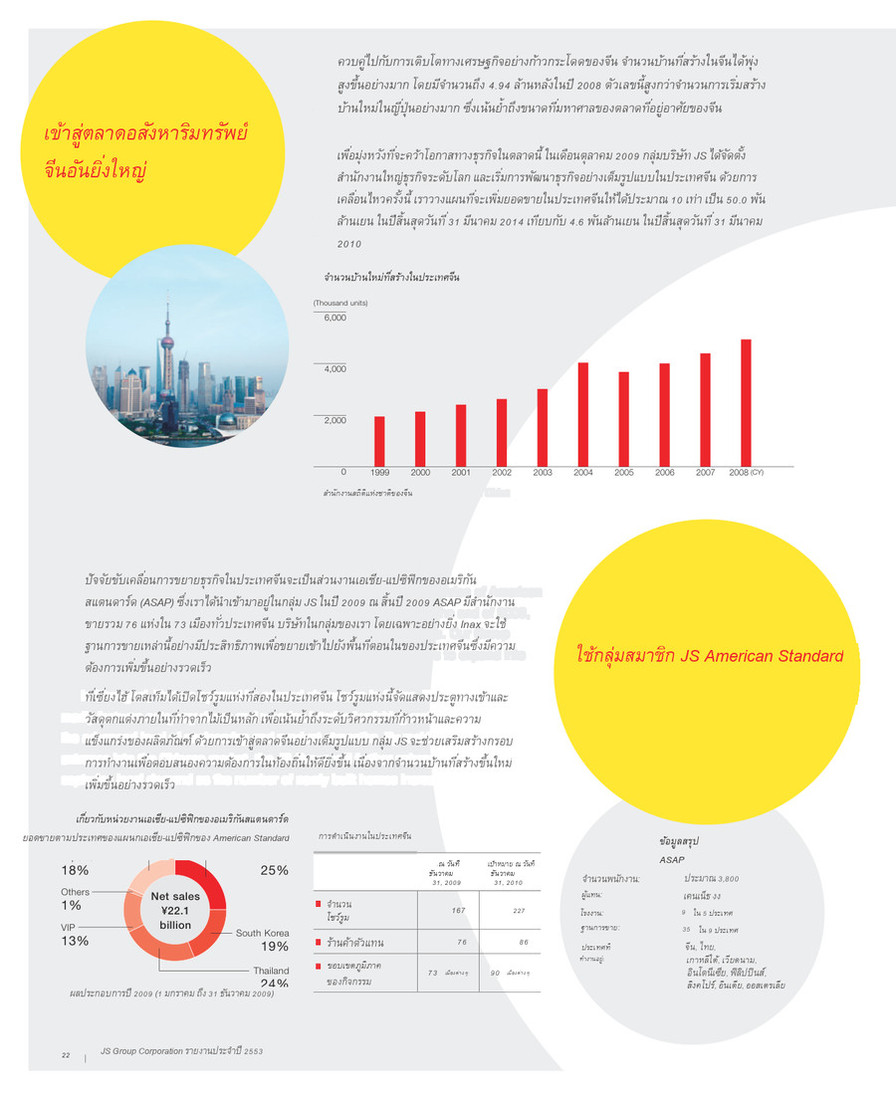} & \dompanel{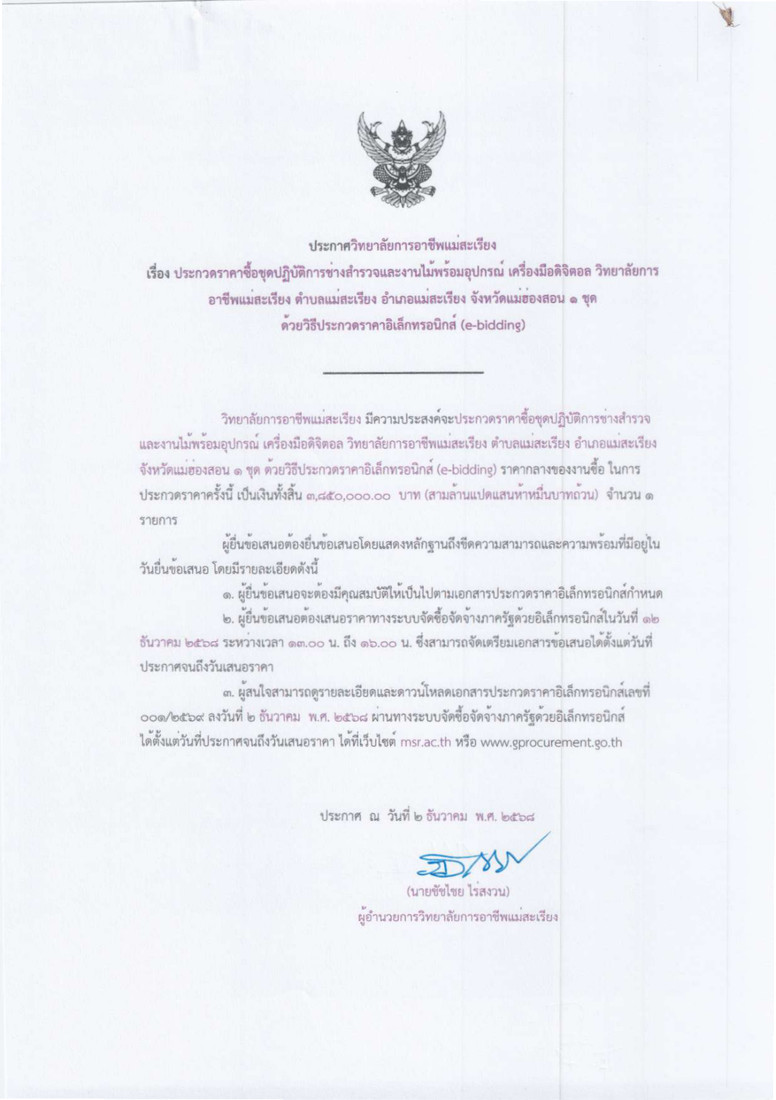} & \dompanel{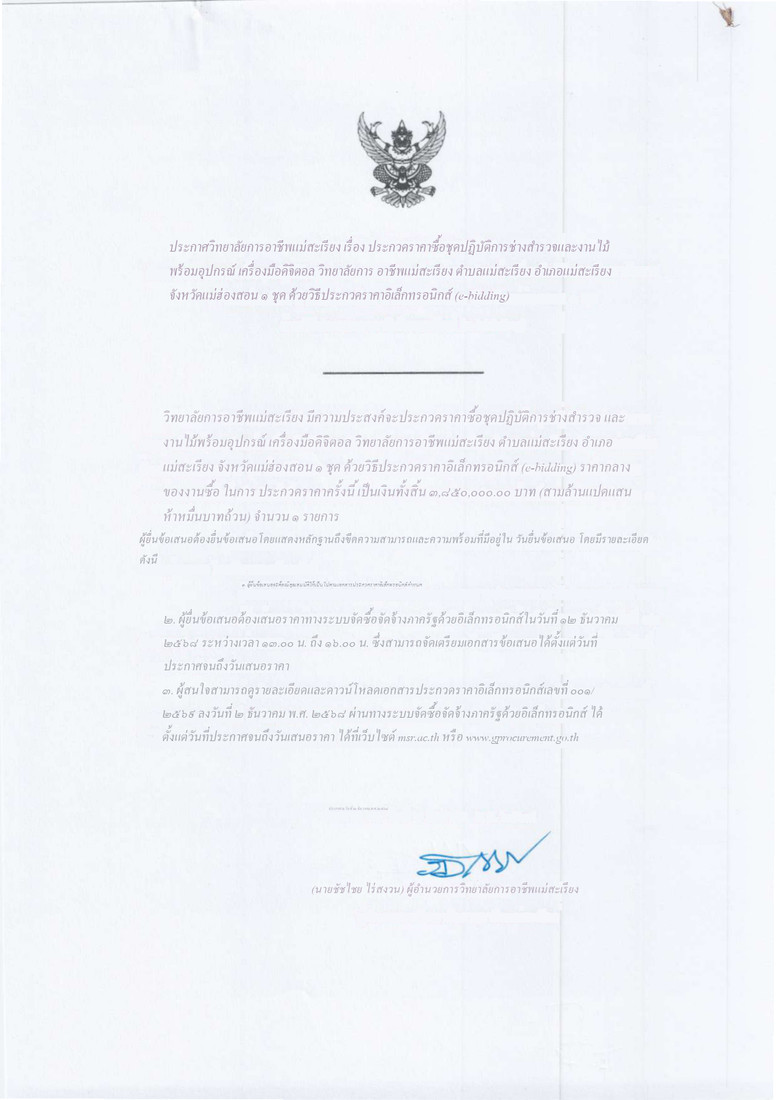} \\
  \end{tabular}
  \caption{The three training sources compared in
  Sections~\ref{sec:in-domain} and~\ref{sec:real-supervision}. \textbf{Out-of-Domain
  Synthetic} (left) translates text from English DocLayNet pages into Thai while retaining
  their layout and non-text pixels. \textbf{Real Thai (Print)} (center) uses each crawled
  Thai page with its original OCR labels. \textbf{In-Domain Synthetic} (right) reconstructs
  the same page by rendering its OCR labels while preserving the non-target page elements.
  The renderer independently selects the
  typeface and size, so its line breaks can differ from the original. The top row is a
  born-digital Comptroller General's Department circular; the bottom row is a scanned
  provincial e-bidding announcement whose scan artifacts and signature remain unchanged.}
  \label{fig:domain-examples}
\end{figure}

Figure~\ref{fig:domain-examples} places the two synthetic sources of
Section~\ref{sec:in-domain} beside the real supervision of Section~\ref{sec:real-supervision}.
Each In-Domain Synthetic example appears beside the Real Thai (Print) page from which it was
reconstructed. The Out-of-Domain Synthetic examples use separate English source documents.
All panels have the same displayed height and retain the aspect ratio of their source pages.

\section{Handwriting Rendering Examples}
\label{app:handwriting-examples}

\begin{figure}[h!]
  \centering
  \setlength{\tabcolsep}{1.5pt}
  \setlength{\fboxsep}{0pt}
  \newcommand{\hwpanel}[1]{%
    \fcolorbox{black!35}{white}{\includegraphics[height=2.24in]{figures/#1}}}
  \newcommand{\hwhead}[2]{%
    \begin{tabular}{@{}c@{}}\footnotesize #1\\[-2pt]\scriptsize\itshape #2\end{tabular}}
  \begin{tabular}{@{}ccc@{}}
    \hwhead{Out-of-Domain Synthetic}{printed typeface} &
    \hwhead{$+$ handwriting typefaces}{rendered outlines} &
    \hwhead{$+$ real glyph instances}{sampled handwriting} \\[2pt]
    \hwpanel{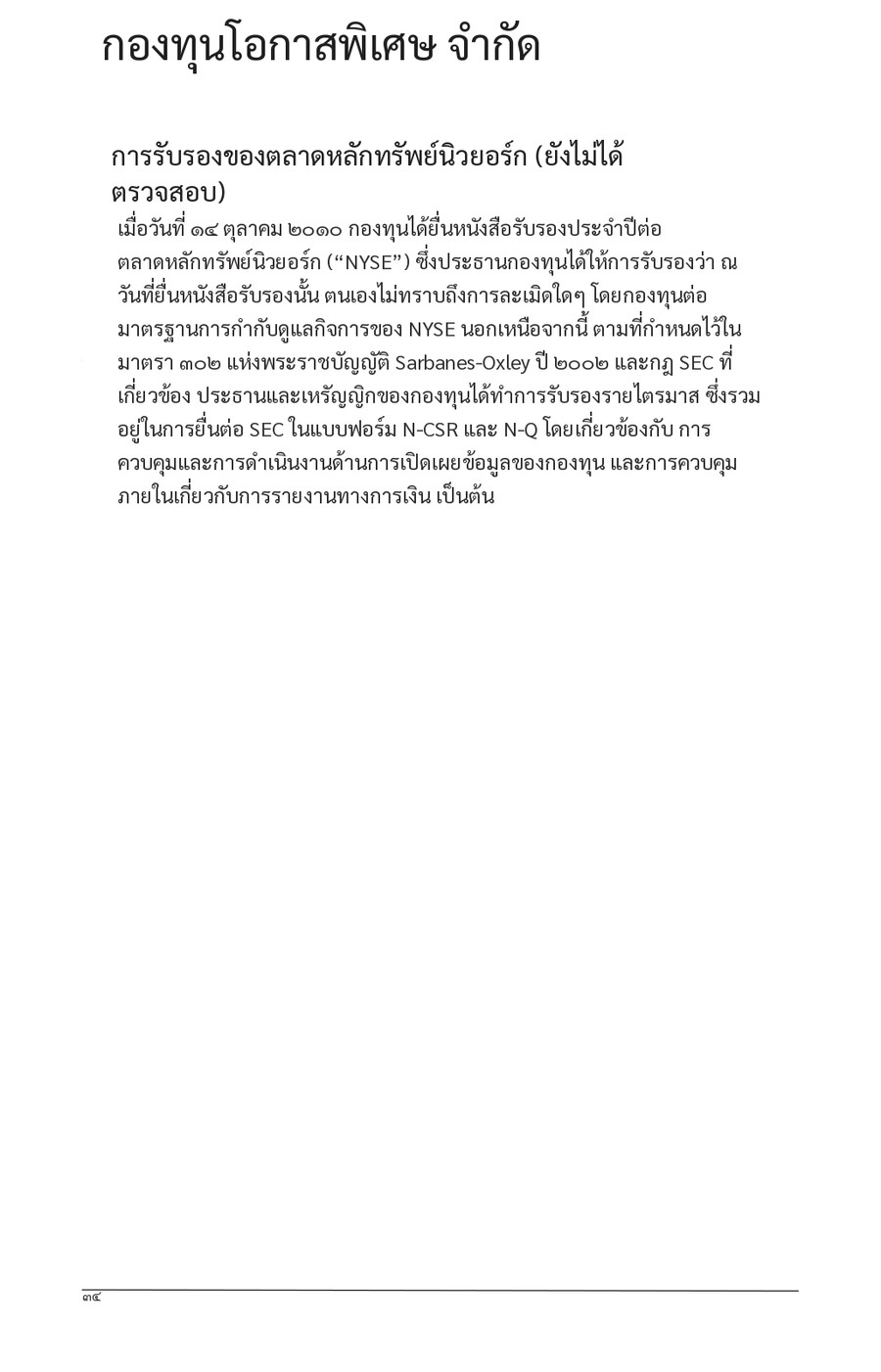} & \hwpanel{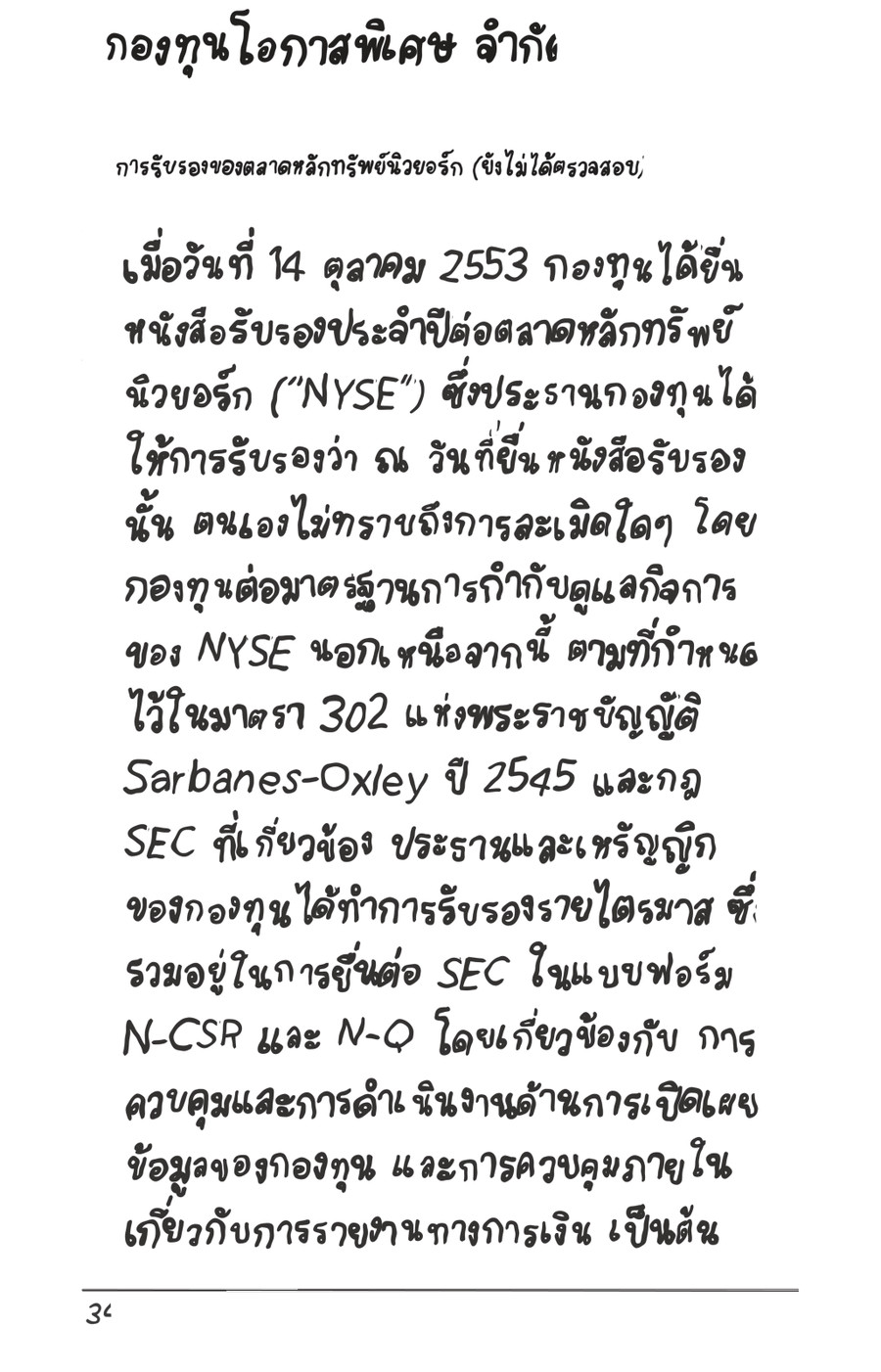} & \hwpanel{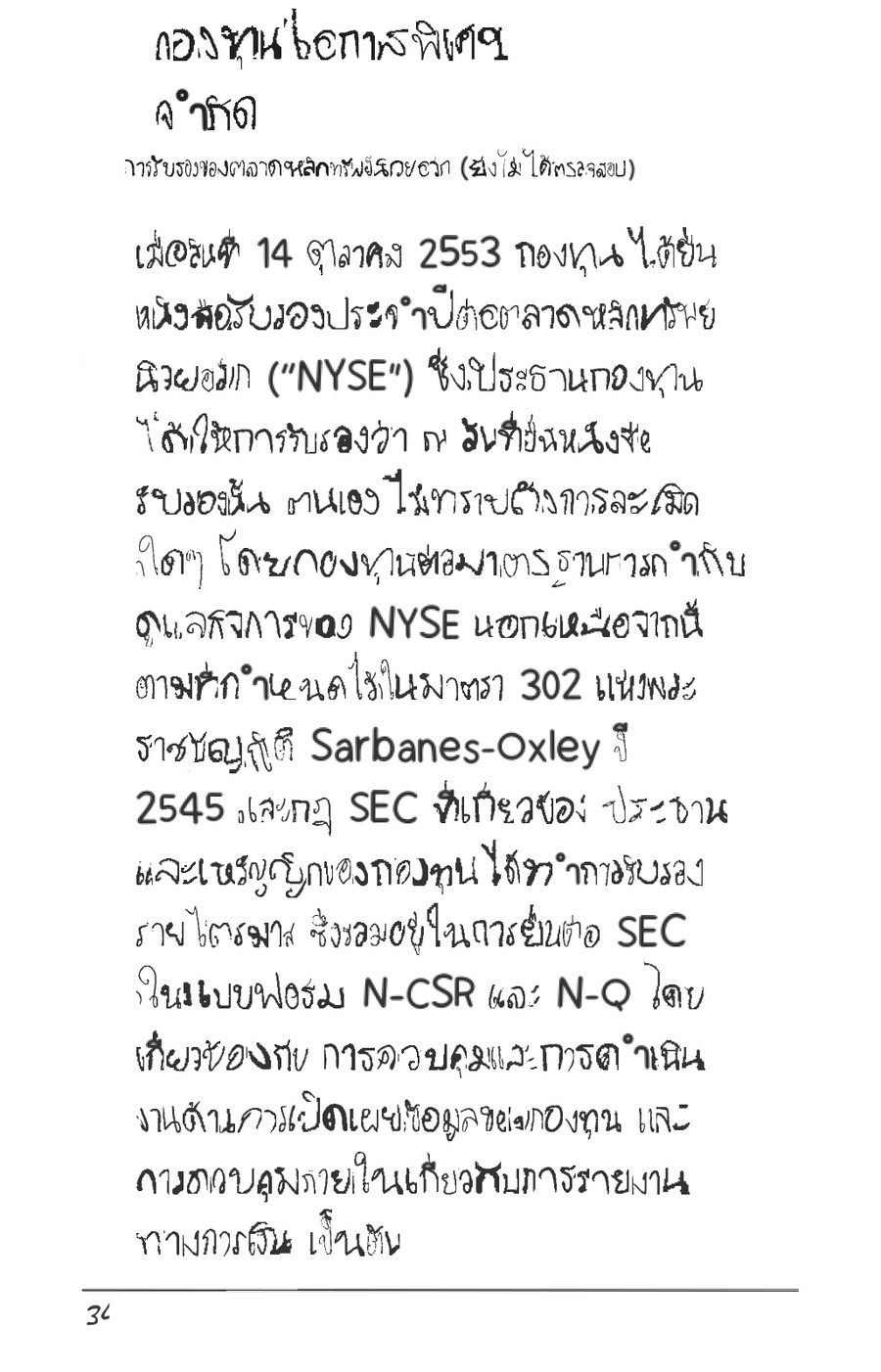} \\[3pt]
    \hwpanel{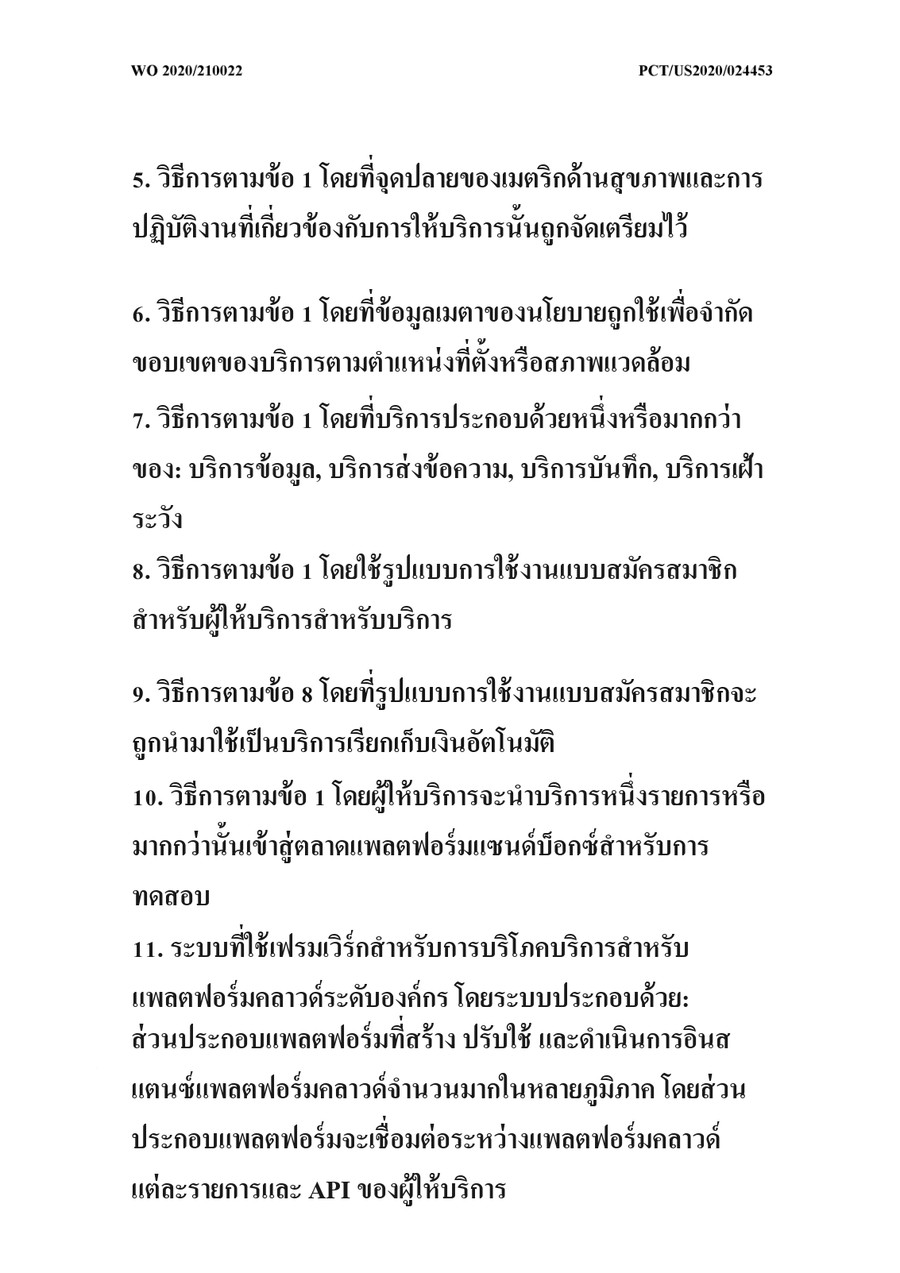} & \hwpanel{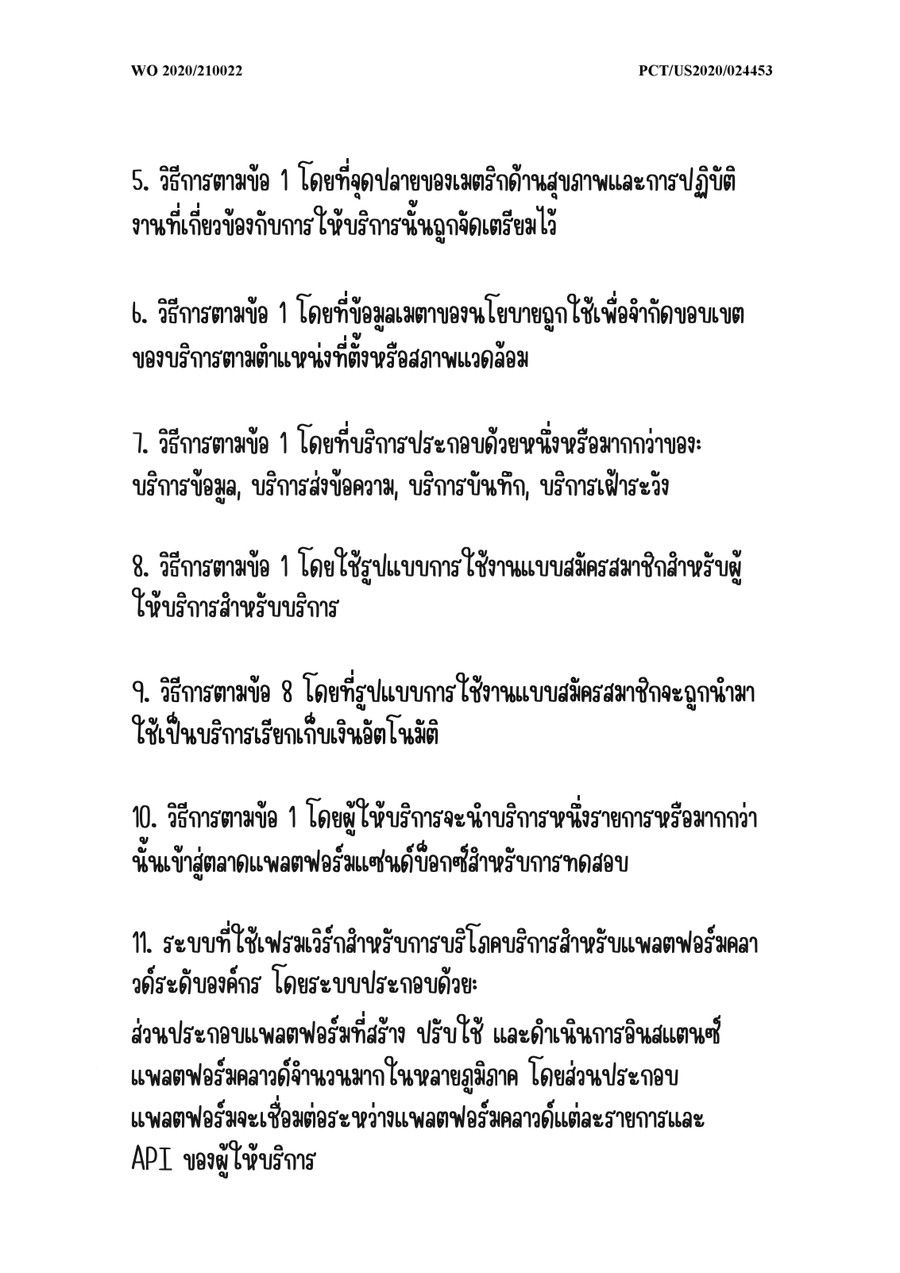} & \hwpanel{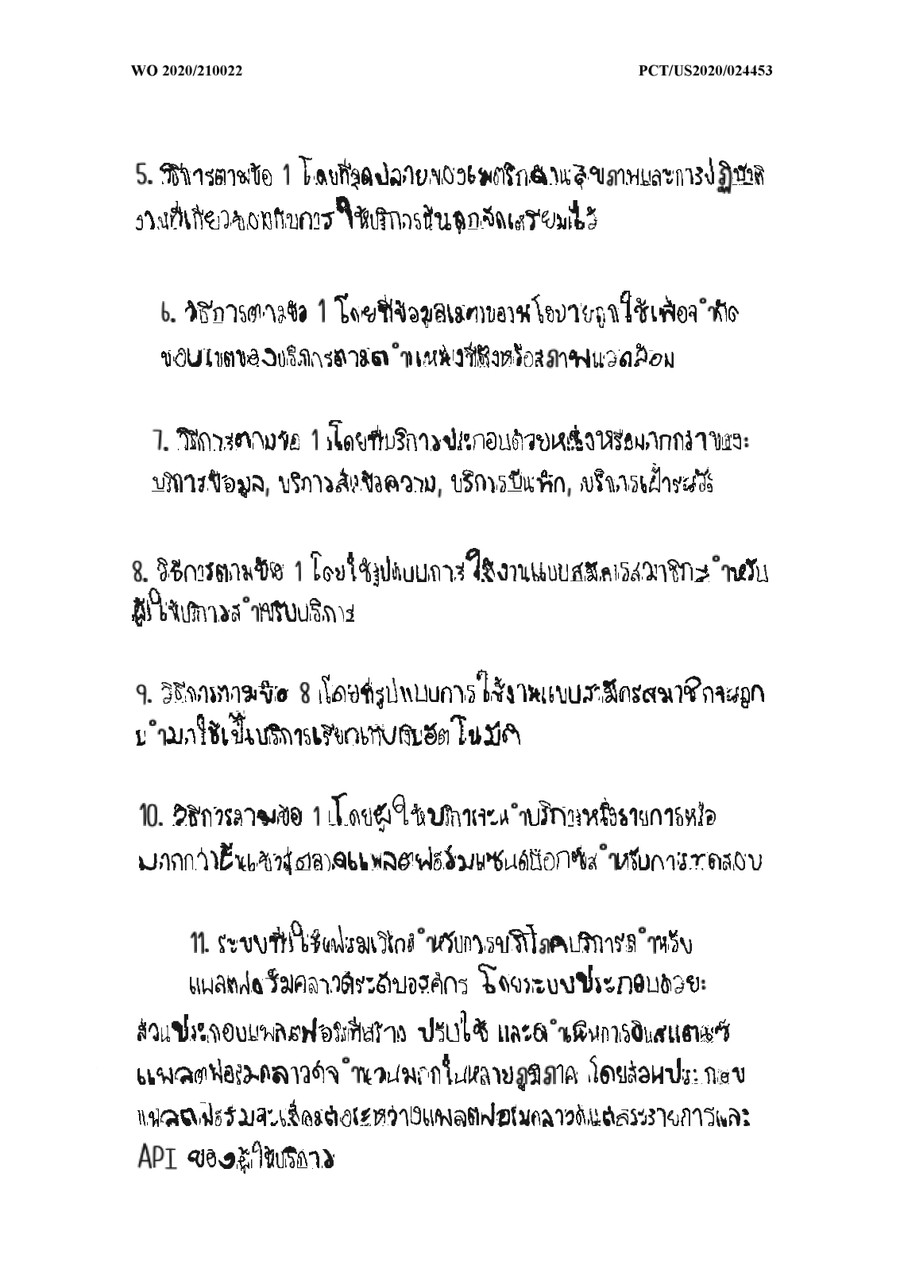} \\
  \end{tabular}
  \caption{Two pages of the handwriting subset under the three conditions of
  Section~\ref{sec:handwriting}. Out-of-Domain Synthetic (left) uses a printed typeface
  sampled from the measured Thai font profile. The handwriting-typeface variant (center)
  changes the sampled face, but repeated characters retain the same outline. The real-glyph
  variant (right) draws each supported character from the glyph bank, introducing
  variation across repeated characters. All three conditions use the same Thai OCR labels
  and region annotations; line breaks differ because the character widths change.}
  \label{fig:handwriting-pages}
\end{figure}

\begin{figure}[h!]
  \centering
  \setlength{\fboxsep}{0pt}
  \newcommand{\hwdet}[1]{%
    \fcolorbox{black!35}{white}{\includegraphics[width=0.97\textwidth]{figures/#1}}}
  \begin{tabular}{@{}l@{}}
    \scriptsize $+$ handwriting typefaces \textit{(rendered outlines)} \\[1pt]
    \hwdet{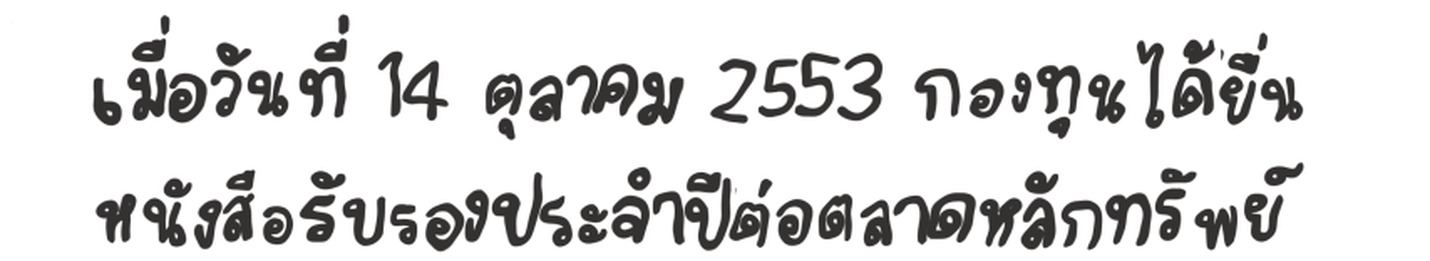} \\[5pt]
    \scriptsize $+$ real glyph instances \textit{(sampled handwriting)} \\[1pt]
    \hwdet{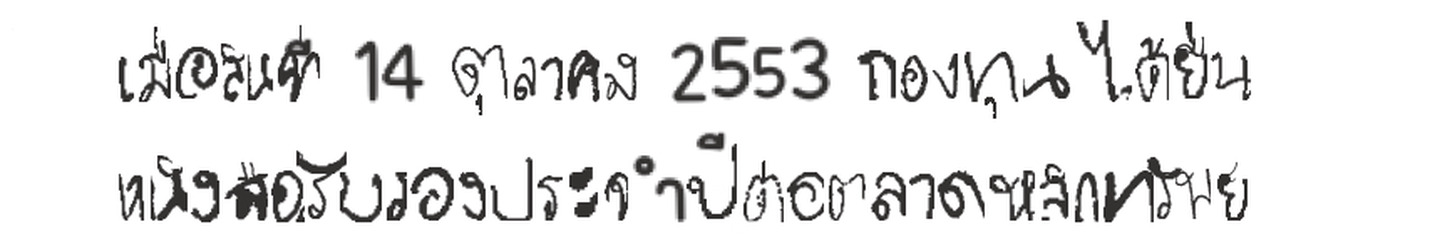} \\
  \end{tabular}
  \caption{The same two lines from the first page of
  Figure~\ref{fig:handwriting-pages}, shown at the same magnification. The typeface variant
  repeats one outline for each character, whereas the real-glyph variant samples separate
  instances with variation in stroke weight, slant, proportion, and local deformation. Half
  of the handwriting-typeface pages also use per-instance stroke distortion, which perturbs
  the rendered outline without changing its underlying shape. Ink height and color remain
  fixed, isolating the source of the strokes.}
  \label{fig:handwriting-detail}
\end{figure}

Figures~\ref{fig:handwriting-pages} and~\ref{fig:handwriting-detail} show the two handwriting
variants from Section~\ref{sec:handwriting}. The real-glyph variant redraws the same 7,131
pages rendered with handwriting typefaces; all other pages remain unchanged. Within the
redrawn pages, 96.3\% of Thai characters use instances from the glyph bank, while the
remaining characters fall back to typeface rendering.

\end{document}